\documentclass[10pt]{article} 
\usepackage[accepted]{tmlr}

\usepackage{amsmath,amsfonts,bm}

\def\eqref#1{equation~\ref{#1}}

\def\1{\bm{1}}

\def\vtheta{{\bm{\theta}}}

\def\vk{{\bm{k}}}

\def\vp{{\bm{p}}}
\def\vq{{\bm{q}}}

\def\vv{{\bm{v}}}

\def\vx{{\bm{x}}}
\def\vy{{\bm{y}}}
\def\vz{{\bm{z}}}

\def\mI{{\bm{I}}}

\def\mK{{\bm{K}}}

\def\mU{{\bm{U}}}
\def\mV{{\bm{V}}}
\def\mW{{\bm{W}}}
\def\mX{{\bm{X}}}

\DeclareMathAlphabet{\mathsfit}{\encodingdefault}{\sfdefault}{m}{sl}
\SetMathAlphabet{\mathsfit}{bold}{\encodingdefault}{\sfdefault}{bx}{n}

\def\gF{{\mathcal{F}}}

\def\gL{{\mathcal{L}}}

\def\sR{{\mathbb{R}}}

\DeclareMathOperator*{\argmax}{arg\,max}
\DeclareMathOperator*{\argmin}{arg\,min}

\usepackage{amsmath,amsfonts}
\usepackage{algorithmic}
\usepackage{algorithm}
\usepackage{array}
\usepackage[caption=false,font=normalsize,labelfont=sf,textfont=sf]{subfig}
\usepackage{textcomp}
\usepackage{stfloats}
\usepackage{url}
\usepackage{verbatim}
\usepackage{graphicx}
\usepackage{hyperref}
\usepackage{cleveref}

\usepackage{epigraph}

\usepackage{multirow}
\usepackage{array}      
\usepackage{booktabs}   
\usepackage{makecell}   
\usepackage{tabularx} 

\title{Beyond the Black Box: A Survey on the Theory and Mechanism of Large Language Models}

\author{\name Zeyu Gan \\
      \name Ruifeng Ren \\
      \name Wei Yao \\
      \addr Gaoling School of Artificial Intelligence, Renmin University of China
      \AND
      \name Xiaolin Hu \\
      \addr Key Laboratory of Multimedia Trusted Perception and Efficient Computing, Xiamen University
      \AND
      \name Gengze Xu \\
      \name Chen Qian \\
      \name Huayi Tang \\
      \name Zixuan Gong \\
      \name Xinhao Yao \\
      \name Pengwei Tang \\
      \name Zhenxing Dou \\
      \name Yong Liu \email liuyonggsai@ruc.edu.cn \\
      \addr Gaoling School of Artificial Intelligence, Renmin University of China}

\begin{document}

\maketitle

\begin{abstract}
The rapid emergence of Large Language Models (LLMs) has precipitated a profound paradigm shift in Artificial Intelligence, delivering monumental engineering successes that increasingly impact modern society. However, a critical paradox persists within the current field: despite the empirical efficacy, our theoretical understanding of LLMs remains disproportionately nascent, forcing these systems to be treated largely as ``black boxes''. To address this theoretical fragmentation, this survey proposes a unified lifecycle-based taxonomy that organizes the research landscape into six distinct stages: Data Preparation, Model Preparation, Training, Alignment, Inference, and Evaluation. 
Within this framework, we provide a systematic review of the foundational theories and internal mechanisms driving LLM performance. Specifically, we analyze core theoretical issues such as the mathematical justification for data mixtures, the representational limits of various architectures, and the optimization dynamics of alignment algorithms. Moving beyond current best practices, we identify critical frontier challenges, including the theoretical limits of synthetic data self-improvement, the mathematical bounds of safety guarantees, and the mechanistic origins of emergent intelligence. By connecting empirical observations with rigorous scientific inquiry, this work provides a structured roadmap for transitioning LLM development from engineering heuristics toward a principled scientific discipline. 
\end{abstract}
\epigraph{\textit{``The grand aim of all science is to cover the greatest number of empirical facts by logical deduction from the smallest number of hypotheses or axioms.''}}{--- Albert Einstein}

\section{Introduction}
\label{sec:introduction}

The recent emergence of Large Language Models (LLMs) has marked a profound paradigm shift in the landscape of Artificial Intelligence (AI). Models such as ChatGPT~\citep{openai2022introducingchatgpt}, DeepSeek~\citep{guo2025deepseek}, Qwen~\citep{bai2023qwen}, Llama~\citep{touvron2023llama}, Gemini~\citep{team2023gemini}, and Claude~\citep{caruccio2024claude} have transcended the boundaries of traditional Natural Language Processing (NLP)~\citep{transformer}, demonstrating capabilities that impact nearly every facet of modern society. As these systems scale, they exhibit behaviors that mimic human-like reasoning~\citep{CoT}, sparking a global transformation in how we interact with information. 

In the history of technological development, engineering triumphs are often inextricably linked to scientific innovation. However, the synchronization between theory and application is rarely instantaneous. 
Consider the trajectory of nuclear physics: from Einstein’s formulation of the mass-energy equivalence equation ($E=mc^2$) in 1905~\citep{einstein1905does} to the detonation of the first atomic bomb at Los Alamos in 1945~\citep{rhodes2012making}, scientists and engineers traversed a forty-year journey to translate theoretical insight into physical reality~\citep{hoddeson1993critical}. 
A similarly extended timeline defines the current era of AI. Approximately 33 years elapsed between the proposal of the Universal Approximation Theorem~\citep{hornik1989multilayer}, which provided the mathematical assurance that neural networks could represent any continuous function, and the emergence of ChatGPT~\citep{openai2022introducingchatgpt}, the definitive proof of that potential. 
Looking ahead from our current vantage point, the quest for Artificial General Intelligence (AGI) necessitates a balanced synergy where continuous theoretical research and rigorous engineering implementation are recognized as equally indispensable pillars. 

\begin{figure*}[tp]
    \centering
    \includegraphics[width=\linewidth]{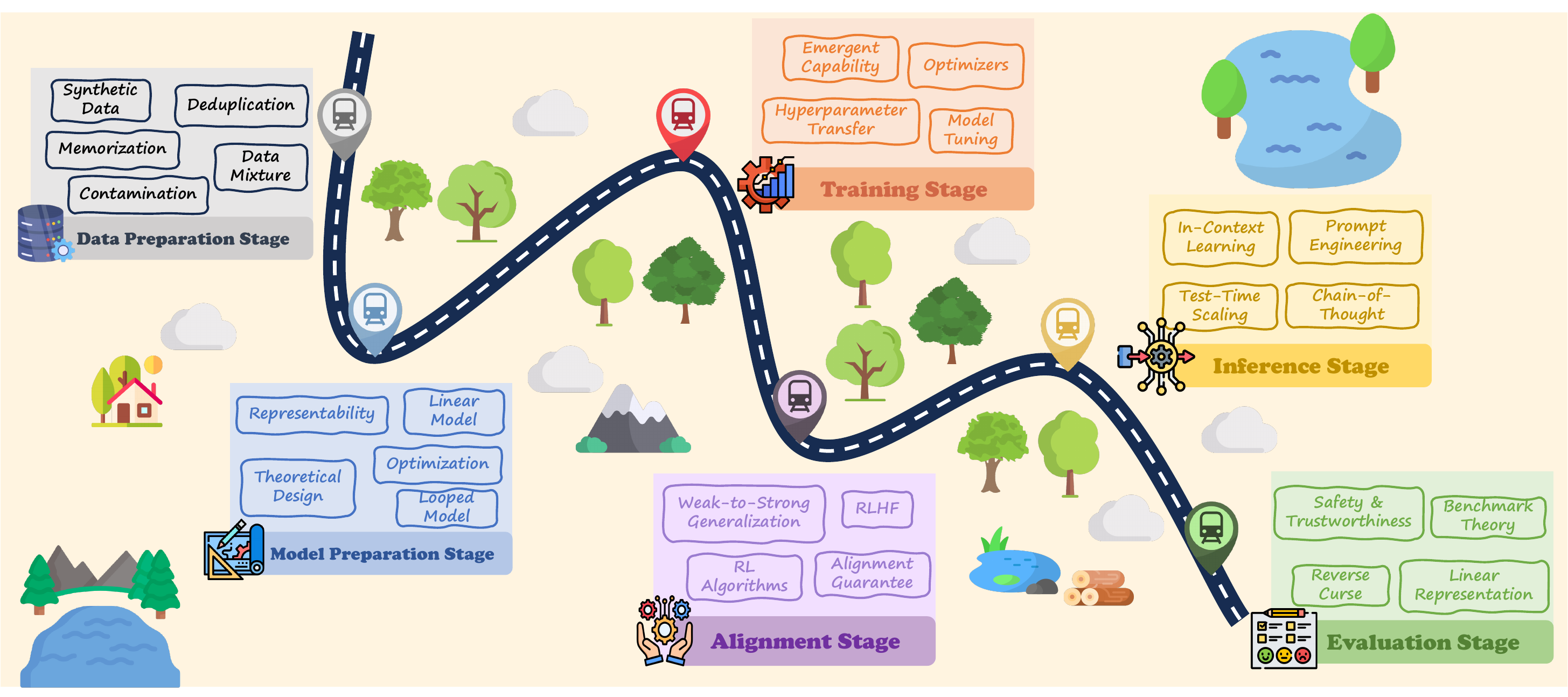}
    \caption{\textbf{The roadmap of LLM theory and mechanisms.} We organize the fragmented theoretical landscape into a unified lifecycle consisting of six stages: Data Preparation, Model Preparation, Training, Alignment, Inference, and Evaluation. The figure visualizes the flow of theoretical inquiry, mapping key sub-topics and algorithmic mechanisms to their respective developmental phases.} 
    \label{fig:intro-pipeline}
\end{figure*}

Throughout these decades, researchers have relentlessly pursued the essence of intelligence through diverse engineering and scientific lenses. At this pivotal moment, with the empirical success of LLMs, we appear closer than ever to unveiling the nature of intelligence. 
Yet, a paradox persists within our current standing: despite the monumental engineering successes of LLMs, our theoretical understanding of them remains disproportionately nascent. While deep learning theory has advanced substantially~\citep{roberts2022principles}, the specific phenomena emerging from LLMs loom like a ``dark cloud'' over the field, shattering previous intuitions and challenging established statistical learning paradigms~\citep{scalinglaw1}. Consequently, we are currently forced to treat LLMs largely as ``black boxes''~\citep{linardatos2020explainable,zhao2024explainability}. They function exceptionally well, yet their internal mechanisms of operation, the how and why behind their efficacy, remain elusive. 

The difficulty in piercing this black box stems primarily from two dimensions. First, the sheer scale of LLMs introduces unprecedented complexity~\citep{scalinglaw1,scalinglaw2}. With parameter counts reaching the trillions and a natural language state space that is combinatorially vast, accurately analyzing the learning dynamics and optimization landscape becomes an arduous mathematical challenge. Second, LLMs exhibit numerous ``emergent'' phenomena that do not appear in smaller models, such as hallucination~\citep{xu2024hallucination}, in-context learning (ICL)~\citep{brown2020language}, scaling laws~\citep{scalinglaw1}, and sudden ``aha moments'' during training~\citep{guo2025deepseek}. These phenomena are difficult to unify under a single theoretical framework, rendering the modeling of LLMs a fragmented endeavor. Consequently, current analyses of LLM theory and mechanisms are often scattered, isolated within specific sub-topics without a holistic view. 

To address this fragmentation, this survey proposes a comprehensive, lifecycle-based perspective. Following the standard LLM pipeline, we categorize the theoretical landscape into six distinct stages as illustrated in~\cref{fig:intro-pipeline}: the \textbf{Data Preparation Stage}, \textbf{Model Preparation Stage}, \textbf{Training Stage}, \textbf{Alignment Stage}, \textbf{Inference Stage}, and \textbf{Evaluation Stage}. By categorizing popular topics and theoretical advances into these stages, we aim to provide a structured roadmap that connects empirical observations with their underlying mechanisms.


The remainder of this paper is structured as follows: \cref{sec:data-preparation-stage} through \cref{sec:evaluation-stage} provide a detailed review of the theory and mechanisms corresponding to each of the six stages, 
and \cref{sec:conclusion} concludes with a discussion on the future of LLM theory. 

\section{Data Preparation Stage}
\label{sec:data-preparation-stage}
The journey of constructing an LLM begins with the data upon which it is built. The \textbf{Data Preparation Stage} encompasses all processes involved in collecting, cleaning, and curating the vast corpora required for training~\citep{radford2019language,albalak2023efficient,bi2025pushing}. This initial stage is arguably the most critical, as the scale, diversity, and quality of the data fundamentally define the limit of a model's potential capabilities, including its knowledge breadth, reasoning abilities, and even its intrinsic biases~\citep{belenki-etal-2025-optimizing,muennighoff2023scaling}. While often perceived as an engineering-heavy process, the choices made during data preparation are deeply intertwined with fundamental theoretical questions about learning, generalization, and information representation. In this section, we review the theory and mechanism of the data preparation stage, from its foundational problems to the theories explaining empirical phenomena, and finally to the open questions that drive future research. 

\subsection{Fundamental Problems}
At its core, the data preparation stage grapples with foundational questions inherited from statistical learning theory and information theory. These problems concern the very nature of the data itself and its theoretical relationship with the learning process, independent of any specific model architecture. Two of the most critical questions are: 

\textbf{(1) How to guarantee better data utilization?} This problem concerns the theoretical relationship between data quality and the learning process. Modern LLM training utilizes rich, heterogeneous, and non-i.i.d. web-scale data. The challenge lies in extending these theories to justify the efficacy of data mixtures and to determine how deduplication and filtering can enhance training efficiency by increasing information density. 

\textbf{(2) How does data affect model performance?} This inquiry seeks to quantify the impact of data characteristics on a model's ultimate capabilities. It involves understanding the trade-off between verbatim memorization and reasoning capabilities, as well as the theoretical limits of synthetic data in recursive self-improvement loops. Furthermore, it addresses how data contamination skews evaluation integrity, forcing a distinction between algorithmic reasoning and the mere recall of benchmark-related samples. 

\begin{figure}[tp]
    \centering
    \includegraphics[width=0.8\linewidth]{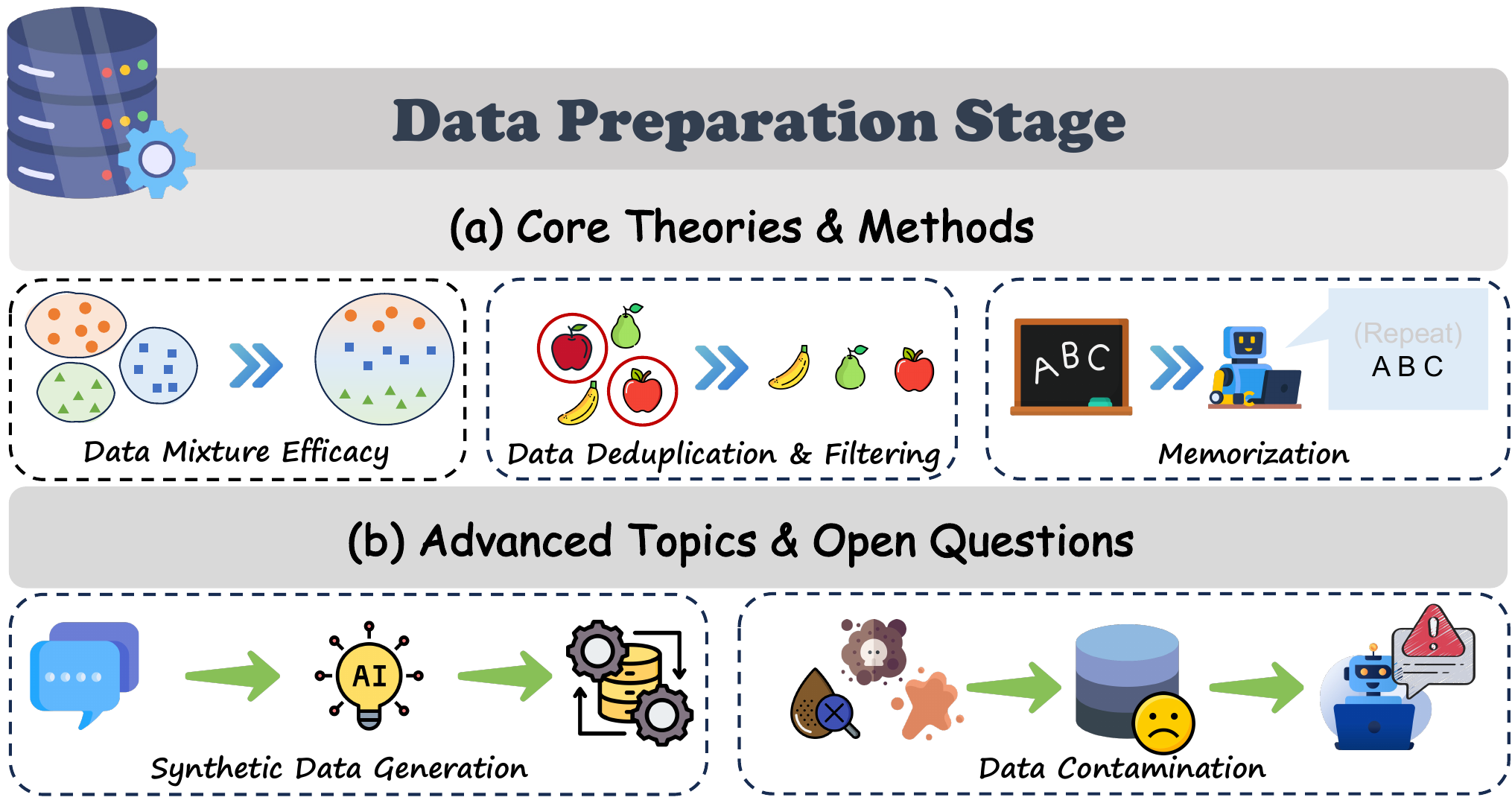}
    \caption{\textbf{An overview of the theoretical landscape in the Data Preparation Stage.} This stage is categorized into two dimensions: \textbf{(a) Core Theories \& Methods} addresses foundational mechanisms including Data Mixture Efficacy (optimizing the proportions of heterogeneous data sources for generalization), Data Deduplication \& Filtering (strategies to enhance training efficiency by dropping redundant data), and Memorization (analyzing the trade-off between verbatim recall and reasoning capabilities). \textbf{(b) Advanced Topics \& Open Questions} highlights frontier challenges, specifically Synthetic Data Generation (investigating the theoretical limits of recursive self-improvement) and Data Contamination (addressing the impact of benchmark leakage on evaluation integrity).} 
    \label{fig:data-preparation-stage}
\end{figure}

These two questions, aiming to improve data quality and understand its impact on model performance, form the bedrock of data-centric LLM theory and mechanism. An illustration of the corresponding topics is shown in \cref{fig:data-preparation-stage}. While classic learning theories provide the fundamental justification for scaling up datasets, their assumption of i.i.d. (independent and identically distributed) samples fails to capture the complex interplay of diversity, source mixture, and quality crucial for modern LLMs. The fundamental problem, therefore, is to extend these theories for the rich, heterogeneous, and non-i.i.d. nature of web-scale text data. To begin tackling this, we subsequently begin to review the core theories and methods in this stage. 

\subsection{Core Theories \& Methods}
These fundamental problems define the ultimate questions in the theory and mechanism of data preparation. To begin answering these profound questions, the academic community has initiated several concrete lines of research, each tackling a specific, empirically observed phenomenon. In what follows, we will review these research efforts, detailing how the study of practical strategies provides valuable insights into our foundational challenges. 

\subsubsection{Data Mixture Efficacy}
A pivotal empirical finding is that performance is not merely a function of data volume, but of its heterogeneity. Models trained on a carefully curated mixture of data from multiple sources (e.g., web text, books, code, scientific articles)~\citep{liu2025rethinking} consistently outperform those trained on monolithic corpora. This observation has spurred a line of research focused on understanding and optimizing the data mixture, which has evolved along three primary axes: theoretical justification, predictive modeling, and algorithmic optimization.

\textbf{Theoretical Foundations for Mixed-Data Training. }
The theoretical analysis is rooted in extensive classic literature on Domain Adaptation~\citep{ben2010theory,mansour2008domain,courty2016optimal}. 
Modern analysis for mixed data training further relies on the perspective of multi-task learning (MTL) or multi-source learning (MSL). 
To explain the strong generalization of highly overparameterized deep models, \citet{zakerinia2025low} propose a modern view based on low intrinsic dimensionality from the MTL perspective. Their key insight is that while a deep model may have a vast number of parameters, its learning process is confined to a low-dimensional manifold. 
Specifically, they provide a key generalization bound (Theorem 2) as follows: 

\begin{equation} \label{eq:MTL-bound}
    \mathcal{R}(f_{1},...,f_{n}) \le \hat{\mathcal{R}}(f_{1},...,f_{n}) + \sqrt{\frac{(l(E)+l_{E}(f_{1},...,f_{n}))\log(2)+\log\frac{1}{\delta}}{2mn}}. 
\end{equation}
In \cref{eq:MTL-bound}, $\mathcal{R}$ represents the true multi-task average risk , and $\hat{\mathcal{R}}$ is the empirical risk on the training data. Their core contribution is that the generalization gap (the square root term) no longer depends on the model's vast number of original parameters, but rather on $l(E)+l_{E}(f_{1},...,f_{n})$, which represents the total compressed encoding length required to jointly encode all $n$ task models (including shared parameters $E$ and task-specific parameters $f_i$). This result rigorously proves theoretically that when multi-task structures are shared (allowing for shorter encoding), the model can achieve stronger generalization, even in the overparameterized state of deep learning. 

Alternatively, \citet{pmlr-v267-wang25eu} offer a pioneering theoretical analysis of MSL within the framework of conditional generative modeling. They establish a general distribution estimation error bound (Theorem 3.2) based on the bracketing number: 

\begin{equation} \label{eq:MSL-bound}
\mathcal{R}_{\overline{TV}}(\hat{p}_{X|Y})\le3\sqrt{\frac{1}{n}(\log\mathcal{N}_{||}(\frac{1}{n};\mathcal{P}_{X|Y},L^{1}(X))+\log\frac{1}{\delta})}. 
\end{equation} 

In \cref{eq:MSL-bound}, the average Total Variation error $\mathcal{R}_{\overline{TV}}$ is controlled by the complexity of the conditional distribution space $\mathcal{P}_{X|Y}$, as measured by its bracketing number $\mathcal{N}_{||}(\cdot)$. Their work formally addresses the question of whether it is more effective to train a single model on all sources or separate models for each one. The authors prove that when source distributions share sufficient ``parametric similarity'' and the model has adequate capacity, multi-source training (which results in a smaller, more constrained distribution space and thus a smaller bracketing number) is guaranteed to achieve a sharper error bound than training on sources in isolation. The theoretical advantage stems from the model's ability to reduce the complexity of the overall distribution space it must learn. 

\textbf{Predictive Models for Data Mixture. }
Besides theoretical validation, researchers have also worked towards creating quantitative models that can predict performance based on the mix. A significant breakthrough was the introduction of ``data mixing laws''~\citep{ye2024data}, which establish a predictable, functional relationship between the mixing proportions of training data and the model's validation loss on each domain. This framework allows for the a priori prediction of a model's overall loss for any given mixture. They propose that the validation loss $L$ on a validation set composed of $K$ (potentially implicit) domains with proportions $s_i$, given training mixture proportions $r_j$ across $M$ domains, can be predicted by: 

\begin{equation} \label{eq:data-mixing-law}
L(r_{1...M})=\sum_{i=1}^{K}s_{i}L_{i}(r_{1...M})=\sum_{i=1}^{K}s_{i}[c_{i}+k_{i}exp(\sum_{j=1}^{M}t_{ij}r_{j})]. 
\end{equation} 
Here, $L_i$ represents the loss on the $i$-th validation domain, and $c_i, k_i, t_{ij}$ are parameters fitted using small-scale experiments. This law enables predicting the performance of large models on unseen data mixtures by fitting the function on results from smaller models and fewer training steps, significantly reducing the cost of mixture optimization. 

This concept was further refined by BiMix~\citep{ge2024bimix}, which proposes a more granular bivariate law. This model explicitly describes how two core variables—the proportion $r_i$ for a specific domain $i$ and the total data volume (represented by training steps $s$)—jointly influence the validation loss $L_i$ on that specific domain: 

\begin{equation}
    L_{i}(r_{i},s)=\frac{A_{i}}{r_{i}^{\alpha_{i}}}(\frac{B_{i}}{s^{\beta_{i}}}+C_{i}), 
\end{equation}

where $A_i, B_i, C_i, \alpha_i, \beta_i$ are fitted parameters for domain $i$. BIMIX provides a per-domain view of how performance scales with both its own proportion and the overall training duration, offering a more detailed predictive model than one based solely on mixture proportions. 

\textbf{Optimization Strategies. }
With predictive models in place, the next logical step is to develop algorithms that automatically find the optimal mixture. Several distinct optimization strategies have emerged based on different theoretical perspectives. For instance, 
REGMIX~\citep{liu2025regmix} frames the problem as a regression task. It operates on the core assumption of ``rank invariance,'' positing that the relative superiority of a data mixture is preserved across different model scales and data volumes. 
Differently, UtiliMax~\citep{held2025optimizing} draws an analogy to financial portfolio optimization. It treats data sources as ``assets'' and seeks a mixing strategy that optimally balances three factors: the ``Utility'' (expected contribution) of each source, ``Diversity'' (to mitigate the risk of overfitting), and ``Scale'' (to avoid over-sampling smaller, high-quality datasets). 
Moreover, DoReMi~\citep{xie2023doremi} leverages min-max optimization, formulating the objective as minimizing the worst-case performance across all data domains, thereby enhancing the model's robustness and uniformity. 
Another particularly
powerful approach, employed by methods like DOGE~\citep{pmlr-v235-fan24e} and ScaleBIO~\citep{pan-etal-2025-scalebio}, is Bilevel Optimization. This framework directly targets generalization by defining a nested objective: the outer loop optimizes the data source sampling weights to minimize validation loss, while the inner loop finds the optimal model parameters that minimize training loss given those weights. 

\subsubsection{Data Deduplication \& Filtering}
Another critical phenomenon is the effectiveness of deduplication. Removing duplicate or near-duplicate examples from the training corpus has become a standard practice, as it has been observed to improve model generalization and reduce verbatim memorization. 

\textbf{The Benefits of Data Deduplication. }
The investigation into deduplication's benefits forms one line of inquiry. \citet{lee-etal-2022-deduplicating} provide a foundational, empirical confirmation of its multiple advantages. This work demonstrates that deduplication directly addresses unnecessary memorization, reducing the frequency of models memorizing training text. 
Alternatively, \citet{kandpal2022deduplicating} propose a core argument: data repetition is the key driver of memorization that leads to privacy risks. This work posits that privacy attacks are successful primarily because of duplicate sequences in the training data. By re-training models on sequence-level deduplicated data, they confirm that this mitigation significantly reduces privacy risks, thus establishing a causal link between data duplication and privacy vulnerabilities. 

\textbf{The Understanding of Deduplication. }
The theoretical understanding of deduplication has evolved significantly from early engineering trade-offs to more sophisticated information-theoretic concepts. Early large-scale datasets, such as The Pile~\citep{gao2020pile}, recognize the importance of deduplication. However, its application was often limited by computational constraints such as only deduplicating within the noisiest subsets rather than globally. This makes deduplication an engineering-heavy compromise rather than a fully realized theoretical application. The primary bottleneck for applying deduplication theory at the trillion-scale was computational. Traditional CPU-based MinHash LSH~\citep{indyk1998approximate} implementations were too slow. This mechanism-level bottleneck was addressed by frameworks like FED~\citep{son2025fed}, which introduces a reusable hash function with lower computational cost and performing end-to-end GPU parallel optimization for the entire MinHash LSH process, reducing tasks from weeks to hours. 
With scalability solved, the theoretical focus shifted. The RefinedWeb~\citep{penedo2023refinedweb} provided a key insight: models trained on aggressively filtered and deduplicated web data could outperform those trained on curated corpora. This suggested that data quality and information density were more critical theoretical levers than simple data source curation. The D4~\citep{tirumala2023d4} framework further evolved this concept. It moves beyond syntactic matching (e.g., hashes) to semantic matching, leveraging pre-trained model embeddings to select a subset of documents that is both de-duplicated and semantically diverse. This mechanism demonstrated tangible performance gains, speeding up training and improving downstream accuracy. The most recent conceptual advance, SoftDedup~\citep{he-etal-2024-softdedup}, addresses a theoretical flaw in ``hard deduplication'' methods: the risk of information loss, and further proposes a soft reweighting mechanism instead. 

\subsubsection{Memorization}
Beyond the practical strategy of deduplication, a core theoretical issue in data preparation is the intrinsic mechanism of memorization~\citep{wei2024memorization}. While often viewed as a privacy risk, memorization is deeply intertwined with the model's learning and generalization capabilities. Research in this area has evolved from observing exact replication to analyzing complex memory representations, quantification methods, and its fundamental trade-offs with generalization. 


\textbf{The Mechanism of Memorization. }
Academic discourse first challenged the traditional view that memorization is caused solely by exact sequence duplication in the training data. A foundational study introduced the concept of ``Mosaic Memory''~\citep{shilov2024mosaic}. This work posits that LLM memorization is not merely verbatim recall, but a more complex process where models can patch together memories by integrating partially overlapping or similar sequences (i.e., fuzzy duplicates) from the training corpus. Building on this, other research redefined memorization from an adversarial perspective, proposing the ``Adversarial Compression Ratio'' (ACR)~\citep{schwarzschild2024rethinking}. The core idea is that a training sequence is considered ``memorized'' if it can be elicited by a prompt that is significantly shorter than the string itself. This metric provides a practical, adversarial view for assessing data usage compliance and potential privacy violations.

\textbf{Quantify the Influence of Memorization. }
Once complex memory forms were defined, the focus shifted to its quantification and prediction. \citet{carlini2022quantifying} confirm that memorization is more prevalent than previously believed and is likely to get worse as models continue to scale, at least without active mitigation. This scaling behavior was specifically quantified for factual knowledge, with one study proposing ``Scaling Laws for Fact Memorization''~\citep{lu-etal-2024-scaling}. It found that a model's fact knowledge capacity exhibits a linear relationship with model size and a negative exponential relationship with training epochs. 
Beyond model scale, data-side characteristics are also a critical factor. The ``Entropy-Memorization Law''~\citep{huang2025entropy} was proposed to investigate the inherent difficulty of memorizing data. This law reveals a linear correlation: the data's entropy is linearly correlated with its memorization score, suggesting that simpler, lower-entropy data is more easily memorized. In terms of predictability, \citet{biderman2023emergent} have further shown that using a partially trained model to predict memorization is more effective than using a small model.

\textbf{Memorization Analysis. }
The ultimate goal of understanding memorization is to differentiate it from generalization and to enable effective control. A key challenge is diagnosing memorization in black-box models. The PEARL~\citep{djire2025memorization} framework was introduced as a novel detection method based on a perturbation sensitivity hypothesis. This hypothesis posits that memorized content is more sensitive to input perturbations, whereas generalized knowledge remains robust. 
Further analysis reveals a clear trade-off between memorization and generalization across different tasks. \citet{wang2025generalization} traced model capabilities back to pretraining data, finding that task dependencies vary significantly. For instance, Factual Question Answering demonstrates the strongest memorization effect, and this effect increases with model size. Conversely, tasks like machine translation and reasoning exhibit greater generalization, tending to produce novel outputs. 
Based on these theoretical insights, researchers have begun exploring active mitigation strategies. ``Memorization Sinks''~\citep{pmlr-v267-ghosal25a} is proposed to activate a unique set of ``memorization neurons'' for each sequence. This mechanism effectively isolates the memorized content, making it easier to remove without compromising general language capabilities, offering a new path to mitigate the negative impacts of memorization. 

\subsection{Advanced Topics \& Open Questions}
\label{app:open-questions-for-data-preparation-stage}
As the field progresses, the focus of data preparation theory is shifting from understanding current best practices to tackling the more profound and forward-looking challenges. These advanced topics explore the theoretical limits and future possibilities of data's role in creating more capable and dynamic AI systems.

\subsubsection{Synthetic Data Generation}
One of the most exciting and debated frontiers is the use of synthetic data and the potential for a self-improvement loop~\citep{villalobos2024position,long2024llms}. Can a model generate new, high-quality data to train its next generation, thereby kicking off a cycle of recursive self-improvement? 
While numerous works adopt synthetic data to improve model training~\citep{patel2024datadreamer,gilardi2023chatgpt,xu2023knowledge,gandhi2024better,liubest}, this idea faces significant theoretical hurdles. A key open question is whether such a process would lead to genuine capability gains or result in model collapse, a degenerative process where the model overfits to its own idiosyncrasies, leading to a gradual loss of diversity and accuracy. Developing a theoretical framework to understand and control the dynamics of this loop is a critical area of research. 


Recent researches have started to establish theoretical frameworks for the utility of synthetic data. \citet{Gan2024TowardsAT} propose a ``reverse-bottleneck'' framework, which posits that a post-trained model's generalization error upper bound is negatively correlated with the ``information gain'' obtained from the generative model. This suggests that so long as the generative model provides sufficient new information, generalization can in principle be improved. Beyond simple augmentation, the nature of the synthetic data is also being explored. For instance, in the domain of mathematical reasoning, \citet{setlur2024rl} find that while fine-tuning on correct synthetic answers offers modest gains, using reinforcement learning on the model's incorrect responses can be twice as sample-efficient. This method helps the model identify and unlearn ``spurious correlations'' (i.e., incorrect intermediate steps that happen to lead to a correct final answer), ultimately scaling the synthetic dataset's efficiency by eight-fold compared to standard positive-only finetuning. 

The primary theoretical hurdle to recursive self-improvement is ``Model Collapse''. \citet{shumailov2023curse} provide a foundational study on this phenomenon, positing that training on generated data leads to an irreversible degenerative process. 
Beyond this, other works have highlighted the limitations of synthetic data in capturing human nuance. \citet{li-etal-2023-synthetic} find that the performance gap between real and synthetic data is smallest for low-subjectivity tasks (like news classification) but much larger for high-subjectivity tasks (like humor or sarcasm detection). This suggests LLMs struggle to generate data with sufficient diversity to capture the complexity of subjective language. This is empirically supported by~\citet{moller-etal-2024-parrot}, which finds that models trained on human-labeled data consistently exhibited superior or comparable performance to those trained on synthetically augmented data. 

In response to the threat of model collapse, various mitigation strategies have emerged. Multiple studies have empirically and theoretically demonstrated that the training workflow is the critical factor~\citep{gerstgrasser2024is,kazdan2024collapse}. A ``replace'' workflow, which discards old data and trains new models only on synthetic data, does lead to collapse. However, an ``accumulate'' workflow, where synthetic data is added alongside the original real data, consistently avoids model collapse and keeps models stable. \citet{seddik2024how} provide a quantitative estimate, concluding that to maintain stability, the amount of synthetic data used must be considerably smaller than the amount of real data in the training mix. This body of work suggests that the value of synthetic data is highly context-dependent: it can improve performance when real data is scarce but may harm it when real data is plentiful. 



\subsubsection{Data Contamination}
Data contamination, the inadvertent inclusion of benchmark evaluation samples within the pre-training corpus, poses another critical open challenge in the data preparation stage. This issue fundamentally undermines the validity of model evaluations, making it difficult to discern true generalization capabilities from mere memorization of seen answers~\citep{deng-etal-2024-unveiling,cheng2025survey,xu2024benchmark}. The theoretical and empirical investigation of contamination can be broadly categorized by its severe impacts and the methods for its detection and mitigation. 

Data contamination is a direct threat to reliable model assessment. 
Studies demonstrate that contamination can drastically skew evaluation scores. \citet{pmlr-v267-kocyigit25a} find that contamination in a machine translation task could severely inflate model capability. This work also revealed that larger models exhibit higher sensitivity to contamination, not greater robustness. \citet{li2024task} further enrich the evaluation, they find that a model's superior performance in apparent zero- or few-shot settings may not stem from genuine generalization but from its exposure to task-related samples during pre-training. The impact on complex reasoning evaluation is particularly stark. \citet{huang2024competition} test models on novel competition problems released after their training data cut-off. They find a ``cliff like decline'' in GPT-4's performance on medium-to-hard problems, strongly suggesting that its high performance on older benchmarks was reliant on memorization rather than genuine algorithmic reasoning. However, contamination is not limited to harmful verbatim copies. \citet{palavalli2024taxonomy} establish through experiments that ``noisy'' or approximate forms of contamination (e.g., masking, augmenting, or noising test examples) can boost performance almost as much as seeing clean, in-domain data. Beyond evaluation, the memorization of contaminated data, especially sensitive information, creates significant privacy vulnerabilities. \citet{zhu2024privauditor} provide a systematic benchmark for assessing these privacy leakage risks, which are exacerbated during model adaptation and fine-tuning. 

In response to these severe impacts, researchers have developed various methods for detection, though effective mitigation remains a significant open problem. \citet{pmlr-v267-choi25b} propose the KDS framework. Instead of simple text matching, it quantifies contamination by measuring the change in the similarity structure of sample embeddings in the model's representation space before and after fine-tuning. \citet{deng2024investigating} introduce an innovative detection method called TS-Guessing. This protocol masks an incorrect answer in a multiple-choice question and prompts the model to fill in the blank. Commercial LLMs were able to ``guess'' the exact missing wrong option with high accuracy, strongly implying they had memorized the full question format. 
Other approaches use targeted queries to excavate a model's memory. \citet{chang2023speak} use ``name cloze'' queries to identify a wide range of memorized copyrighted books, which in turn contaminated downstream evaluation tasks. Similarly, \citet{dodge2021documenting} confirm that the C4 dataset~\citep{10.5555/3455716.3455856} contains contaminated examples from NLP benchmarks. 
Contamination is not just a pre-training issue. \citet{tao2025detecting} address the challenge of detection after RLHF, where optimization erases traditional likelihood signals. The proposed ``Self-Critique'' method probes for ``policy collapse'' by comparing the token-level entropy sequences of an initial response and a second, alternative critique response, where high similarity indicates memorization.

\section{Model Preparation Stage}\label{sec:model}
Once the foundational dataset is prepared, the focus shifts to the vessel of learning itself. The \textbf{Model Preparation Stage} encompasses the critical decisions regarding the model's blueprint, including the selection of its core architecture, the design of the tokenization scheme, and the strategy for parameter initialization. This architectural foundation is paramount, as it dictates the model's inductive biases, its scaling properties, and the very landscape of the optimization problem to be solved. While many architectural choices are guided by empirical breakthroughs, they are deeply rooted in theoretical questions about computational efficiency, information flow, and the representation of complex patterns. 

\subsection{Fundamental Problems}

After data preparation, a key question arises: how to choose an \textbf{appropriate}, \textbf{powerful} and \textbf{efficient} model architecture.
In fact, the design and selection of a suitable deep learning model architecture are not only related to the latent characteristics of the training data being handled, but are also influenced by the training paradigm adopted such as next-token prediction (NTP) or masked language modeling (MLM).
However, as self-supervised learning paradigms have become increasingly popular especially after the success of large general-purpose language models, the backbone architectures of such models can often be conveniently transferred to different modalities and training settings. 
A typical example is the powerful attention-based Transformer architecture.
Therefore, in this section, we focus on the theoretical analysis of mainstream architectures that can serve as (potentially) general-purpose model frameworks, while deliberately leaving aside discussions specific to particular modalities or training paradigms that rely on intricate design details.

This section mainly revolves around the following core questions (topics):

\textbf{(1) How to theoretically evaluate the ``power" of a model architecture?} 
    This problem focuses on the rigorous analysis of a model's intrinsic properties, specifically its representability: the capacity to solve or approximate given classes of functions. The challenge is to determine the theoretical limits of architectures under realistic constraints such as finite precision, width, and depth. This involves establishing upper bounds on the model size required to realize specific capabilities and lower bounds that characterize the minimum circuit or communication complexity needed to solve computational tasks. 
	
\textbf{(2) How to theoretically understand and guide the design of model architectures?} 
    This inquiry aims to interpret the internal operations of well-performing models through formal frameworks to inspire principled improvements. A primary challenge is relating the forward pass of stacked architectures to unrolled optimization processes, where each layer is viewed as an iterative step toward minimizing a latent objective function. This includes understanding how model structures arise from principles of information compression, energy-based models, or test-time training paradigms. Furthermore, it addresses the ``no free lunch'' trade-off between sub-quadratic computational efficiency and representational bottlenecks in linear and recurrent models. 

\begin{figure}[tp]
    \centering
    \includegraphics[width=0.8\linewidth]{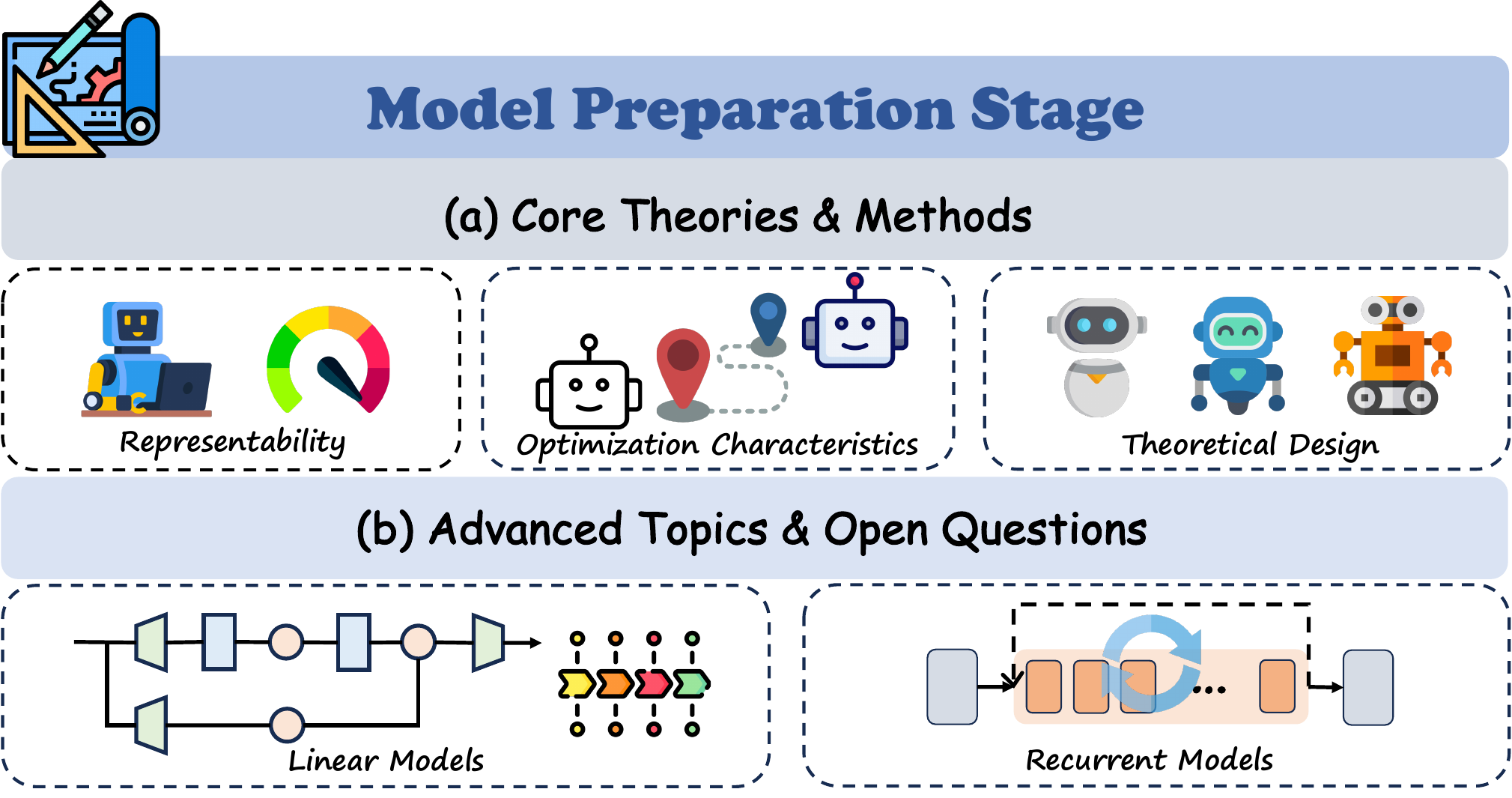}
    \caption{\textbf{An overview of the theoretical landscape in the Model Preparation Stage.} This stage is categorized into two dimensions: \textbf{(a) Core Theories \& Methods} addresses foundational principles including Representability (analyzing expressive power and fundamental limits), Optimization Characteristics (investigating training dynamics and properties), and Theoretical Design (interpreting internal operations through formal frameworks). \textbf{(b) Advanced Topics \& Open Questions} highlights frontier challenges, specifically Linear Models (addressing the efficiency-representation trade-off) and Recurrent Models (exploring weight-tied architectures for iterative reasoning).} 
    \label{fig:model-preparation-stage}
\end{figure}

Of course, discussions on these questions often overlap. For example, according to the ``no free lunch" principle, there is usually a trade-off between model performance and computational cost, that is, efficient models may come with potential limitations in representational power. Moreover, the theoretical understanding of a model’s architecture is closely related to the observable characteristics it exhibits. 
We present a visualization of the topics in this stage in \cref{fig:model-preparation-stage}. In what follows, we start with the core theories and methods relevant to model design. 

\subsection{Core Theories \& Methods}
In this section, we delve into the two core questions outlined above, aiming to highlight the community’s remarkable theoretical efforts in uncovering the capabilities and underlying mechanisms of model architectures.
We first begin with an examination of representational capacity, which theoretically explains what kinds of problems a model can or cannot solve. 
This topic mainly concerns the ideal potential of a model—its ability to provide solutions in principle—without addressing whether the model can actually reach those solutions through training.
Therefore, we next turn our attention to the training dynamics of models, investigating how they behave and what properties they exhibit during the learning process.
Finally, we focus on a topic more closely aligned with practical applications, that is, we try to understanding what models are doing internally from a theoretical standpoint and how such insights can guide the design of more effective and practically usable architectures.

\subsubsection{Representability of Models}\label{sec:model-represent}
A model's expressive power or representability refers to whether it is capable of representing or solving a given class of functions or problems. 
Although the mere existence of such solutions does not guarantee that standard training procedures will discover them, expressive power is still crucial: it reveals the fundamental limits of what a model can do, independent of optimization or data issues.

The community has renewed interest in the expressive capacity of Transformers, especially in terms of \textbf{universal approximation} \citep{yun2019transformers, jiang2024approximation, kajitsuka2023transformers}.
\citet{yun2019transformers} show that for any sequence-to-sequence function, there exists a Transformer that can approximate it, where the number of layers scales exponentially in model dimension $d$ or input sequence length $T$ and the size of each layer is independent of $d$ and $T$.
\citet{jiang2024approximation} derive explicit Jackson-type approximation bounds for Transformers by introducing new complexity measures to construct appropriate approximation spaces and their results show that Transformers approximate efficiently when the temporal dependencies of the target function exhibit a low-rank structure.
\citet{kajitsuka2023transformers} demonstrate that once the feed-forward layer quantizes continuous inputs, even a one-layer, single-head self-attention module becomes a universal approximator for continuous permutation equivariant functions on a compact domain. 
More recently, \citet{su2026sparsity,su2026variational} also extend the representability analysis towards more complex mixture-of-experts (MoE) architectures by introducing tropical geometry. 

In addition, there are also many works that study representability through \textbf{Turing completeness}, asking whether a model can effectively simulate each step of a Turing machine—and thus inherit the full computational power of Turing-computable problems \citep{dehghani2018universal, perez2021attention, wei2022statistically}.
\citet{perez2021attention} prove that Transformers are indeed Turing complete under the assumption of infinite precision.
\citet{dehghani2018universal} show that standard finite-precision Transformers are not Turing complete and the proposed Universal Transformer which combines parallel self-attention with recurrence can overcome this limitation.
\citet{wei2022statistically} demonstrate that Transformers can statistically meaningfully approximate Turing machines running in time $O(T)$, with sample complexity polynomial in the alphabet size, state-space size, and $\log(T)$.
In fact, beyond pursuing idealized theoretical analyses, there is growing interest in understanding the theoretical limits of Transformers under more realistic constraints—namely finite precision, width, and depth. Such analyses typically fall into two categories: upper bounds and lower bounds on the model’s expressive power, which we elaborate on below.
For more detailed discussions and broader context, we refer to the recent survey literature \citep{represent-survey}.

\textbf{Upper bounds.}
Upper bounds typically rely on specific constructions to demonstrate that a model can represent a given function or solve a given task, i.e., that a valid solution indeed exists. 
Once such a solution is found, it provides an upper limit on the model size required to realize this capability \citep{hahn2020theoretical, bhattamishra2020ability, yao2021self, wen2023transformers, liu2022transformers}.
Because these results often rely on constructive proofs, the tasks or problems they address are typically case-by-case.
Although self-attention networks cannot process formal languages with hierarchical structure \citep{hahn2020theoretical, bhattamishra2020ability}, such as $\rm{Dyck}_k$, \citet{yao2021self} 
demonstrate that they can process $\rm{Dyck}_{k,D}$, the subset of $\rm{Dyck}_k$ with depth bounded by $D$.
\citet{wen2023transformers} show that different attention patterns can be Learned to generate bounded $\rm{Dyck}$ and interpretability via local (``myopic") analysis can be provably misleading on Transformers.
\citet{liu2022transformers} prove that a shallow Transformer with $o(T)$ layers can exactly simulate any finite-state automaton processing a sequence of length $T$ and $O(\log T)$-depth Transformers always exist to simulate any automaton of length $T$.
Moreover, even $O(1)$-depth solutions are surprisingly common.

\textbf{Lower bounds.}
Lower bounds show that any model capable of effectively representing a certain function or solving a certain task must have a size at least as large as some specified threshold.
A useful approach is to view the model as a circuit, allowing its expressive power to be characterized using circuit complexity, which studies the minimum circuit size or depth required to solve a computational problem \citep{merrill2023parallelism, li2024chain, hao2022formal}.
\citet{merrill2023parallelism} show that Transformers with $O(1)$-depth and log-precision can only solve problems within the class $TC^0$.
If the precision is further restricted to be constant, then such Transformers are limited to solving problems in $AC^0$ as shown by \citet{li2024chain}.
\citet{hao2022formal} prove that generalized UHAT (unique hard attention) models can only recognize $AC^0$ languages while an averaging hard-attention (AHAT) model that can recognize non-$AC^0$ languages.
Another line of work approaches the question through the lens of communication complexity. 
These studies recast the anchoring problem into a known communication problem and then identify the resulting communication bottlenecks imposed by model width \citep{sanford2023representational, sanford2024transformers, peng2024limitations, chen2024theoretical, chen2024theoretical}. 
\citet{sanford2023representational} introduce the “sparse averaging” task and show that Transformers achieve only $O(\log T)$ communication complexity, in contrast to the polynomial requirements of RNNs and feed-forward networks. 
Furthermore, \citet{sanford2024transformers} demonstrate that log-depth Transformers can solve some basic computational tasks that cannot be efficiently handled by other neural sequence models or by sub-quadratic Transformer approximations.
\citet{peng2024limitations} employ communication complexity to show that a Transformer layer cannot reliably compose functions once the function domains become sufficiently large, a limitation that has been linked to the emergence of hallucinations.
\citet{chen2024theoretical} further provide the first unconditional lower bound for multi-layer decoder-only Transformers: for any fixed depth $L$, an $L$-layer Transformer must have polynomial width in order to perform the sequential composition of $L$ functions over $T$ tokens.
In addition, by reducing the in-context learning problem to a set-disjointness task, \citet{arora2024just} demonstrate that a recurrent model’s ability to recall information is sensitive to the order in which inputs are presented.

\subsubsection{Optimization Characteristics of Models}
The analysis of optimization dynamics in large language models primarily includes the following three aspects: optimization analysis of transformers under/without the in-context learning (ICL) mechanism, as well as the analysis of loss landscapes. 

\textbf{Optimization Analysis of Transformers under ICL Regimes. }
For the first aspect, \citet{Shen2025} theoretically investigate the training dynamics of a single-layer Transformer model for in-context classification tasks on Gaussian mixtures, demonstrating that under optimization via gradient descent, the model can converge to the global optimum at a linear rate. 
\citet{Zhang2024} study the training dynamics of a Transformer with a single linear attention layer during in-context learning for linear regression tasks, showing that the model can find the global minimum of the objective function. 
\citet{Chen2024unveiling} use gradient flow to analyze how a simplified Transformer architecture with two attention layers performs ICL, revealing the collaborative mechanism of its components. 
\citet{Gong2025disentangling} analyze the optimization dynamics of a single-layer Transformer with normalized ReLU self-attention under ICL mechanisms, indicating that smaller eigenvalues preserve basic knowledge, while larger eigenvalues of attention weights capture specialized knowledge. 
\citet{Zheng2024} examines whether autoregressively trained Transformers implement ICL by learning a meta-optimizer. They demonstrate that under specific conditions on the initial data distribution, the trained Transformer indeed learns to perform one-step gradient descent to solve ordinary least squares (OLS) problems in-context. 
\citet{Chen2024training} prove that the training dynamics consist of three phases: warm-up, emergence, and convergence, with ICL capabilities rapidly emerging during the emergence phase. 
\citet{Huang2024} further extend previous analyses from linear attention to softmax attention, demonstrating that on balanced data, the model converges to near-zero prediction error through a two-phase process. On imbalanced data, the model exhibits ``staged'' convergence. 
\citet{Kim2024} theoretically study how Transformers with both MLP and attention layers learn nonlinear features in in-context learning (ICL). Under the assumption that the attention layers converge rapidly, the authors show that the infinite-dimensional loss landscape for MLP parameters exhibits a benign non-convex structure. 

\textbf{Optimization Analysis of Transformers without ICL Regimes. }
The aforementioned studies mainly focus on demystifying the mechanism of ICL. Apart from them, there are also studies that directly study the training behaviors of transformers without ICL. 
\citet{Nichani2025} demonstrate that a single-layer Transformer with self-attention and MLP can achieve perfect prediction accuracy when the number of self-attention parameters or MLP parameters scales almost linearly with the number of facts. 
\citet{Tian2023} 
reveal that the self-attention mechanism exhibits a ``scan and snap'' dynamic: initially distributing attention uniformly across all tokens, it gradually focuses on ``distinctive'' tokens that are discriminative for predicting specific next tokens while reducing attention on ``common'' tokens that frequently appear across different next-token predictions. 
\citet{Ren2024} analyze the training dynamics of a single-layer Transformer on a synthetic dataset, showing that the optimization process consists of a ``sample-intensive'' stage and ``sample-efficient'' stage. 
\citet{Pan2025} propose a mathematical framework based on compression theory to explain the behavior of LLMs. They conceptualize LLM training as first learning and compressing common syntactic patterns, then progressively acquiring and storing knowledge from common to rare. 

\textbf{Optimization Landscape. }
In addition to directly analyzing the optimization dynamics of transformers, some studies have also explored the model's optimization process from the perspective of loss landscapes. 
\citet{Wen2025} propose the ``River Valley Loss Landscape'' hypothesis to analyze the effectiveness of the Warmup-Stable-Decay (WSD) learning rate schedule. 
Through theoretical analysis, they demonstrate that during the stable learning rate phase, a higher learning rate causes parameters to oscillate significantly between the ``hillsides'', while also enabling faster progress along the direction of the ``river'' at the bottom. In the decay phase, the rapidly decreasing learning rate reduces oscillation amplitude, allowing parameters to move closer to the ``river'', leading to a swift decrease in loss. 
Similarly, \citet{Liu2025}, drawing from principles of classical thermodynamics, argue that the model training process can be decomposed into two dynamical stages: a fast dynamics phase characterized by rapid oscillations between hillsides, and a slow dynamics phase involving gradual drift along the river direction. 
\citet{Gong2025what} further extend the ``River Valley Loss Landscape'' framework by proposing two types of river valleys: U-shaped and V-shaped valleys. U-shaped valleys are wide and flat, where optimization tends to stagnate. In contrast, V-shaped valleys feature narrow bottoms and steep sides, allowing parameters to ``jump'' between valley walls while progressing along the river direction. 

\subsubsection{Theoretical Design of Models}\label{sec:model-design}

While analyzing the representational capacity and optimization properties of existing high-performing architectures is undoubtedly fascinating, a crucial prerequisite is that such architectures must first exist for analysis. Although the design of most mainstream and well-performing architectures today still largely depends on engineering intuition and empirical experience, the research community has been making concerted efforts to interpret the underlying mechanisms of model architectures from a theoretical perspective. Building upon these insights, researchers aim to design new architectures that may prove to be both practically useful and conceptually enlightening.

\textbf{Unrolled optimization perspective.} A mainstream and widely appreciated principle is to relate the forward pass of the stacked layers to an unrolled optimization process \citep{ISTA, unroll-dictionary, hinton2022forward, monga2021algorithm, zhang2018ista, guo2023contranorm}. 
Given the input $\vz$ and some model $f$ with $L$ layers, the core idea of the unrolled optimization is to interpret the layer-wise computation of the model as performing an iterative optimization on some latent objective function $F$, that is,
\begin{equation}
\begin{aligned}
    &\vz^* = \argmin_{\vz} F(\vz)  \Longleftrightarrow f: f: \vx = \vz^0 \cdots\to \vz^{l-1} \xrightarrow{f^l}\vz^{l}\to\cdots \vz^L = \vz^* \\
    \end{aligned}
\end{equation}
where $f^l$ is the $l$-the layer and $\vz^l$ is its corresponding output. 
In other words, each layer of the model can be viewed as a single step of an optimization algorithm seeking to minimize or optimize the underlying objective.
Given the success of the Transformer architecture, it is natural for researchers to attempt to understand its structure from principled perspectives \citep{whiteTF, whiteTF-JMLR, whiteTF2, IB-TF, attention-only-TF, yang2022transformers, ren2025transformers}.
One prominent direction is to relate the Transformer’s design to an objective function associated with information compression.
\citet{whiteTF} show that Transformer-like deep network layers can naturally be connected to an optimization process aimed at sparse rate reduction.
More specifically, given the input data $\mX \in \sR^{d\times N}$, they denote $(\mU_k)_{k=1}^{K}$ to be the set of bases of the mixture of low-dimensional $K$ Gaussian distributions.
Then the objective function $F$ can be formalized as
\begin{equation}
\begin{aligned}
	\argmax_{\boldsymbol{Z}=f(\boldsymbol{X})} F &= \mathbb{E}_{Z}\left[\Delta R(\boldsymbol{Z};\boldsymbol{U}_{[K]})-\lambda\|\boldsymbol{Z}\|_{0}\right] \\
    &=\mathbb{E}_{\boldsymbol{Z}}\left[R(\boldsymbol{Z})-R^{c}(\boldsymbol{Z};\boldsymbol{U}_{[K]})-\lambda\|\boldsymbol{Z}\|_{0}\right], 
    \end{aligned}
\end{equation}
where $R$ and $R^c$ are estimates of lossy coding rates \citep{ma2007segmentation, yu2020learning, redunet}.
The objective aims to maximize the information gain for the final token representations by maximizing $\Delta R$ while promoting the sparsity by minimizing the $\ell^{0}$ norm.
It is worth noting that the optimization of $\Delta R$ shares the same underlying inspiration as the design of ReduNet \citep{redunet}. However, here the optimization of $R^c$ ultimately gives rise to the multi-head attention structure, whereas the remaining optimization of $R(\boldsymbol{Z}) - \lambda\|\boldsymbol{Z}\|_{0}$ corresponds to a structure analogous to a feed-forward network (FFN) \citep{ISTA}.

In addition to the interpretation based on rate reduction, \citet{IB-TF} approach the emerging visual grouping phenomenon observed in Vision Transformers from the perspective of the information bottleneck. 
They showed that the iterative solution to the information bottleneck objective can be expressed in the form of self-attention.
\citet{attention-only-TF} point out that compressing noisy token representations and the corresponding denoising operations can naturally give rise to the form of multi-head self-attention.
Other works relate the optimization objective $F$ to energy-based principles \citep{HN-is-all-you-need, EnergyTF, hu2023sparse, wu2023stanhop, ren2025transformers, hu2025hyper}.

\textbf{Test-time Training Perspective.} Although Transformers have achieved widespread success across tasks in different modalities, their quadratic complexity with respect to sequence length often becomes unacceptable under resource-constrained conditions. In addition to improving the efficiency of the Transformer architecture itself, researchers have also begun to focus on designing more efficient model architectures, among which an important line of work is called \textbf{test-time training (regression)} in \citet{TTT, yang2023gated, von2025mesanet, wang2025test, behrouz2024titans, behrouz2025s}.

Generally, the design of this framework can be roughly divided into two stages \citep{wang2025test}. 
The first stage uses a function $f_t$ to store memory in a regression manner at $t$-th test step, and the second stage uses this function for retrieval. 
Formally, similar to attention mechanisms, we transform the token $\vz$ into the form of a query $\vq$, and convert the previous $t$ interacted tokens $\vx$ into key-value pairs $(\vk_1, \vv_1), (\vk_2, \vv_2), \dots, (\vk_t, \vv_t)$.
Then, the output $\vy_t$ at $t$-th step can be formalized as 
\begin{equation}
	\begin{aligned}
		{\rm Memorization:~~} &f_t = \argmin_{f \in \gF} \sum_{i=1}^t \gamma_i \left\| \vv_i - f(\vk_i) \right\|^2,  \\
		{\rm Retrieval:~~} &\vy_t = f_t(\vq),
	\end{aligned}
\end{equation} 
where $\gamma_i$  controls the importance of each association.
When we modify the regression objective including the weighting factor $\gamma_i$, the family of functions $\gF$ and the optimization algorithm, we can derive most existing forms of linear attention.

For the most basic linear attention, we can assign all weights $\gamma_i$ to 1 equally, consider the function $f_t$ in the linear function $\gL_{\rm Linear} = \{f | f(\vk) = \mW \vk, \mW \in \sR^{d_v \times d_k} \}$, and use the analytical solution from Newton’s method, that is, $\mW_t = \mV_t^T \mK_t (\mK_t^T \mK_t)^{-1}$ where $t\geq d_k$.
Then after applying the approximation $(\mK_t^T \mK_t)^{-1} \approx \mI$, we can obtain the simplest form of linear attention formalized as 
\begin{equation}
\begin{aligned}
\vy_t &= f_{\rm Linear}(\vq_t) = \mV_t^T \mK_t (\mK_t^T \mK_t)^{-1} \vq_t \\ 
&\approx \mV_t^T \mK_t  \vq_t = \sum_{i=1}^t \vv_i \vk_i^T \vq_t.
\end{aligned}
\end{equation}
In fact, when we extend the above case using the kernel trick, where a kernel feature map $\phi$ is used to strengthen the representation of $\vk_t$ and $\vq_t$ \citep{elu+1, performer}.
This leads to the unnormalized softmax attention, which is also referred to as the dual model by \citet{ren2024towards}.
In addition, when we assign different weights in the linear-attention setting, this gives rise to the family of gated linear attentions such as RetNet \citep{RetNet, yang2023gated, LRU, RWKV}. 
When the weight $\gamma_i$ depends on the input, it is often interpreted as a forget gate. 
Existing studies also show that state-space models can be viewed as a branch of gated linear attention \citep{Mamba2, Mamba_thu_linear_attention, ren2024exploring}.

Furthermore, when we consider online or streaming setting and apply different gradient-descent algorithms, we obtain the existing family of linear models known as fast weight programmers and online learners \citep{schmidhuber1992learning, DeltaNet, longhorn, GatedDeltanet}.
More specifically, if we perform single-example SGD at each time step and initialize $\mW$ using the linear mapping obtained in the previous step, we can get the form of Delta Rule \citep{DeltaNet}, that is, $\mW_t = \mW_{t-1}(\mI - \beta_t \vk_t \vk_t^T) + \beta_t \vv_t \vk_t^T$ where $\beta_t$ is the learning rate at time $t$.
Longhorn \citep{longhorn} extends the above form by using the adaptive step sizes.
In addition, Gated DeltaNet \citep{GatedDeltanet} combines DeltaNet \citep{DeltaNet} with the forget-gate mechanism of Mamba-2 \citep{Mamba2}, which can be viewed as adding an $F$-norm regularization on $\mW$ to the original objective and performing single-example SGD.

\subsection{Advanced Topics \& Open Questions}
\label{app:open-questions-for-model-preparation-stage}
Although Transformers have become the dominant architecture for modern large language models, the community's pursuit of more powerful and efficient model designs has never ceased. 
This section approaches this frontier topic from two perspectives: linear models and recurrent models, encompassing both theoretical considerations and practical explorations.

\subsubsection{Linear Models \& No free Lunch}
As discussed in Section \ref{sec:model-design}, despite the remarkable performance of Transformers across a wide range of tasks, their quadratic computational cost remains a significant obstacle to broad deployment in real-world settings \citep{transformer}. 
This has motivated a surge of interest in more efficient architectures whose computational and memory costs scale linearly with sequence length, including RetNet \citep{RetNet}, RWKV \citep{RWKV}, gated linear attention \citep{yang2023gated}, TTT\citep{TTT}, Mamba\citep{Mamba, Mamba2}, Longhorn \citep{longhorn}, gated DeltaNet \citep{DeltaNet, GatedDeltanet}.
These models are now widely recognized as belonging to the family of linear RNNs or as instances of the test-time training paradigm \citep{yang2023gated, wang2025test}.
However, the well-known \textbf{``no free lunch"} principle quickly comes into play: linear models often gain efficiency at the expense of representational power.
Intuitively, as such models must compress past information into a fixed-size state without knowing what future inputs will be, two inherent difficulties arise. First, a constant-size state cannot scale with sequence length, causing substantial information loss on long inputs. 
Second, if future patterns deviate from the prior encoded in this compression rule, the compressed representation may completely fail.
These limitations are reflected in recent theoretical findings. 
\citet{phonebook} show that Transformers can copy sequences of exponential length, whereas fixed-state models are fundamentally limited by their finite memory.
Similarly, \citet{Lvkaifeng_rnn_not_tf} demonstrate that generalized RNNs even equipped with chain-of-thought reasoning cannot perform associative recall or other tasks requiring precise contextual retrieval unless they are augmented with retrieval-augmented generation (RAG) or followed by a Transformer layer.

Even so, this does not imply that RNNs are necessarily weaker than Transformers.
For example, \citet{bhattamishra2024separations} prove that bounded Dyck languages can be recognized by constant-size RNNs, while a single-layer Transformer requires linear width. 
\citet{merrill_mamba} further show that the expressive power of linear RNNs with diagonal transition matrices is comparable to that of Transformers (both lying within $TC^0$), yet allowing data-dependent non-diagonal transitions enables linear RNNs to surpass $TC^0$ class \citep{merrill_mamba, grazzi2024unlocking, siems2025deltaproduct}.
These observations point toward an appealing research direction: \textbf{Hybrid architectures} that combine linear models with Transformers.
\citet{Lvkaifeng_rnn_not_tf} theoretically show that simply introducing just one single Transformer layer into RNN is sufficient to enhance its in-context retrieval capability and close the representation gap with Transformers. 
Practical evidence also suggests that hybrid architectures, such as combining Mamba with Transformers, can achieve high efficiency while keeping comparable performance \citep{Mamba_large_ex, Jamba, Zamba}. 
In addition, recent work has explored incorporating the Delta Rule into Transformers to further strengthen their expressive power \citep{zhong2025understanding, deltaformer}.

\subsubsection{Recurrent Models \& Looped Transformers}
Beyond the pursuit of more efficient linear models, recurrent architectures have also begun to re-enter the spotlight in the community. This renewed interest is driven by several factors. 
On one hand, the emergence of chain-of-thought (CoT) reasoning has dramatically boosted models’ expressive power \citep{CoT, hedi_CoT, malach2023auto, merrill2023expressive, li2024chain}, prompting researchers to consider how such iterative reasoning capabilities might be implicitly baked into the model’s inductive bias \citep{yu2025enhancing}. 
More broadly, a strengthened understanding of scaling laws has highlighted that performance gains come not only from scaling data and model size during training \citep{scalinglaw1, scalinglaw2}, but also from increasing test-time computation \citep{test-time-scalinglaw}, for example, by allowing the model to perform recurrent or iterative reasoning \citep{geiping2025scaling, zhu2025scaling, wu2025parallel}.
In fact, the study of recurrent or weight-tied architectures has a long and rich history, providing a strong foundation for these recent developments.

As discussed in Section \ref{sec:model-represent}, \citet{dehghani2018universal} introduce the Universal Transformer, which improves generalization by sharing parameters across layers and allowing the model to flexibly adjust its iterative depth. 
\citet{giannou2023looped} further propose treating Transformers as programmable computational units, where a fixed layer is repeatedly applied to execute instructions encoded in the input sequence. 
\citet{looptf-yangliu} incorporate the looping paradigm directly into the Transformer's iterative computation process, enabling the model to more effectively learn tasks that require internal learning algorithms.
\citet{gatmiry2024can} study whether looped Transformers can implement multi-step gradient descent in an in-context learning setting. 
\citet{looptf-fan, looptf-hedi} demonstrate that looped Transformers achieve substantially better length generalization compared to fixed-depth Transformers.
\citet{saunshi2025reasoning} demonstrate that many reasoning problems require greater depth rather than more parameters, and that looped models can achieve more effective reasoning while using significantly fewer parameters.
Collectively, these studies highlight the advantages of recurrence primarily through theoretical analyses or small-scale experiments.
More recently, \citet{geiping2025scaling} use recurrence as a prior for implicit reasoning in latent space, scaling the model to 3.5B parameters and showing performance competitive with non-looped models tens of billions of parameters in size. 
Similarly, \citet{loopTF-scaling} introduce Ouro, a family of pre-trained looped language models scaling up to 2.6B parameters and trained on 7.7T tokens. 
\citet{loopTF-parallel} propose the Parallel Loop Transformer (PLT) architecture to improve computational efficiency when leveraging recurrence.
\citet{MoR} develope Mixture-of-Recursions (MoR), which combines parameter sharing with adaptive computation to unlock stronger model performance.

\section{Training Stage}
\label{sec:training-stage}
With both the foundational dataset prepared and the model's architectural blueprint finalized, the journey moves to the computationally intensive heart of LLM creation: the \textbf{Training Stage}. This unified stage encompasses the entire learning process, transforming the static architecture into a potent and practical artifact. 
The stage commences with Pre-Training, a massive-scale, self-supervised process where the model ingests the prepared corpus, typically by optimizing a next-token prediction objective. This is where the model's foundational capabilities are forged, imbuing it with vast linguistic knowledge, factual information, and nascent reasoning abilities. Following this, the model undergoes Supervised Fine-Tuning (SFT), the first step in adapting it to human intent. Here, the pre-trained model is further trained on a smaller, high-quality dataset of labeled instruction-response pairs, adapting its general predictive capabilities to specific conversational and task-oriented formats. 

\subsection{Fundamental Problems}
The Training Stage transforms the static, initialized architecture into a potent and practical artifact through two critical phases: massive-scale pre-training and subsequent task-oriented supervised fine-tuning. This entire process is governed by fundamental theoretical questions concerning how learning occurs at an unprecedented scale and how that learned knowledge can be effectively adapted. The core theoretical challenges in this stage can be distilled into two fundamental problems: 

\textbf{(1) How do simple learning objectives forge complex, emergent capabilities at scale?} The dominant paradigm, pre-training, relies on a remarkably simple self-supervised objective, such as next-token prediction. Yet, this process imbues the model with vast linguistic knowledge, factual information, and nascent reasoning abilities. A central problem is to move beyond empirical observation and develop a theoretical framework that explains this emergence. This involves understanding the precise relationship between scale (data, parameters, compute) and capability, which is the core inquiry of Scaling Laws , and probing the mechanisms that form the Origin of Intelligence  from a simple predictive loss. 

\textbf{(2) What are the principles of effective and efficient knowledge adaptation?} A pre-trained model is a general-purpose artifact, not yet optimized for human intent. The second fundamental problem is understanding how to adapt this model. This requires a theoretical grasp of the Fine-Tuning process: How do we instill new, specific knowledge (e.g., instruction following) without catastrophically forgetting the model's general capabilities? Furthermore, given the immense size of these models, how can this adaptation be achieved efficiently? This question drives the theoretical and practical development of Parameter-Efficient Fine-Tuning (PEFT)  methods, which seek to optimize a small subset of parameters while preserving, or even enhancing, the model's foundational knowledge.

\begin{figure}[tp]
    \centering
    \includegraphics[width=0.8\linewidth]{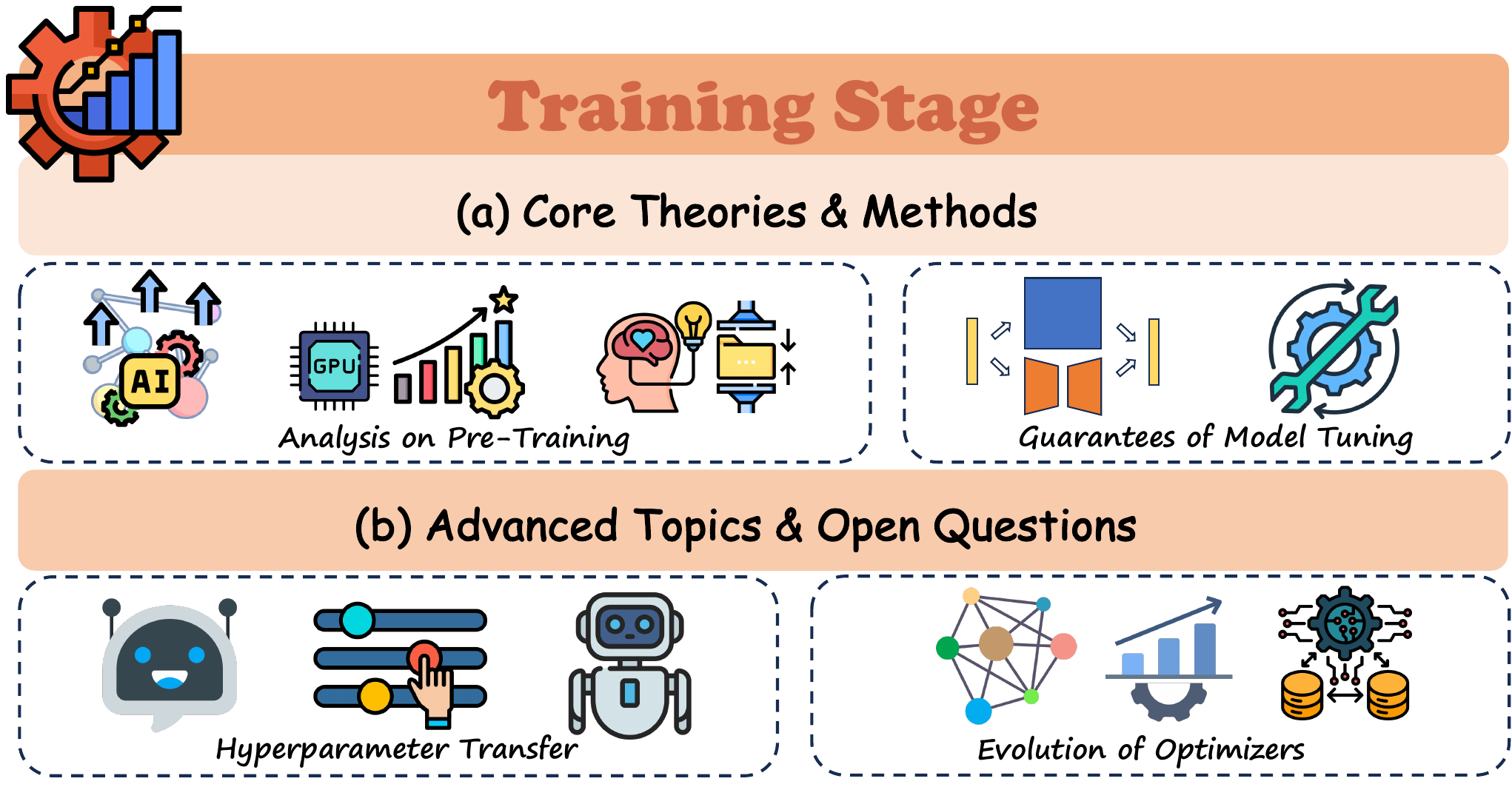}
    \caption{\textbf{An overview of the theoretical landscape in the Training Stage.} This stage is categorized into two dimensions: \textbf{(a) Core Theories \& Methods} addresses mechanisms of knowledge acquisition, including Analysis on Pre-Training (foundations of knowledge acquisition and scaling laws) and Guarantees of Model Tuning (mechanisms and optimization of fine-tuning paradigms). \textbf{(b) Advanced Topics \& Open Questions} highlights frontier challenges, specifically Hyperparameter Transfer (zero-shot transfer of configurations across scales) and Evolution of Optimizers (matrix-aware and adaptive methods for LLMs).} 
    \label{fig:training-stage}
\end{figure}

These two questions, which concern the creation of foundational knowledge via pre-training and the adaptation of that knowledge via fine-tuning, form the theoretical bedrock of the Training Stage. A detailed illustration of the corresponding topics is shown in \cref{fig:training-stage}. In what follows, we review the core theories and methods the community has developed to address these profound challenges. 

\subsection{Core Theories \& Methods}
The fundamental problems characterize the essence of the training stage. In the real-world LLM pipeline, the training stage is further divided into pre-training and fine-tuning processes, which have different goals though share similar training paradigms. In what follows, we will review the theoretical advancements from both aspects. 

\subsubsection{Analysis on Pre-Training}
The pre-training phase is where the model's foundational capabilities are forged. This massive-scale, self-supervised process imbues the model with vast linguistic knowledge, factual information, and nascent reasoning abilities. The theoretical inquiry in this area focuses on some primary axes: understanding why and how the knowledge learned during pre-training is beneficial for downstream tasks, and formalizing the relationship between scale (data, parameters, compute) and capability, commonly known as Scaling Laws~\citep{kaplan2020scaling}. Building upon these, the community further seek for the explanation for the intelligence of LLM. 

\textbf{The Benefits of Pre-Training. }
A significant body of theoretical work seeks to explain why self-supervised pre-training is so effective for transfer learning. Initial research provided a direct mathematical link, proving that a language model achieving $\epsilon$-optimal cross-entropy loss during pre-training can, in turn, enable a simple linear classifier to achieve an error rate of $\mathcal{O}(\sqrt{\epsilon})$ on downstream natural classification tasks~\citep{saunshi2020mathematical}. 
Subsequent work posits that pre-training enables the model to capture the underlying latent variable information within the text data~\citep{wei2021pretrained}. This analysis also helps explain the efficacy of different adaptation methods, noting that prompt-tuning requires weaker non-degeneration conditions than head-tuning, providing a theoretical basis for its strong performance in few-shot settings. 
The quality of the pre-training task itself is also a critical factor. \citet{zhao2023blessing} provide a statistical theory demonstrating that high class diversity in the pre-training objective is key to improving the sample efficiency of downstream tasks. 

More general frameworks have been proposed to formalize the entire transfer process. \citet{deng2024generalization} derive a generalization bound for the fine-tuned model, revealing that its performance depends on four key factors: Representation Transferrability, Representation-induced Rademacher Complexity, Domain Heterogeneity, and the generalization ability of the pre-training task itself. A unified perspective is further offered to theorize that pre-training learns the ``contexture''—the top-$d$ singular functions of the association between inputs and their contexts~\citep{pmlr-v267-zhai25c}. A representation that successfully learns this contexture is proven to be optimal for downstream tasks that are compatible with that context. 
To avoid the costly process of fine-tuning to find the best model, \citet{zhang2025assessing} introduce the DISCO framework. It uses Singular Value Decomposition (SVD) to analyze a model's features, operating on the insight that different spectral components of the features have different degrees of transferability. 

\textbf{Scaling Laws. }
Scaling laws are a set of empirical and theoretical principles that describe the predictable, power-law relationship between a model's performance and increases in scale. 
A foundational work~\citep{hoffmann2022training} in this area establish that for compute-optimal training, model size ($N$) and the amount of training data ($D$) should be scaled proportionally with the compute budget ($C$), specifically $N_{opt} \propto C^{0.5}$ and $D_{opt} \propto C^{0.5}$. This finding reveals that many previous large-scale models were significantly undertrained. 
However, these laws can be disrupted. \citet{dohmatob2024tale} provide a theoretical framework explaining that as synthetic, AI-generated data enters the training corpus, it can alter or break traditional scaling laws, leading to performance degradation and model collapse. 

Given the high cost of training, new methods for studying these laws have emerged. \citet{ruan2024observational} propose a method that analyzes public models to bypass costly retraining, finding it can accurately predict complex performance changes, including phenomena previously considered ``emergent''. 
The nature of emergence itself is also being explained by scaling. \citet{wu2025ushaped} suggest emergence is not a mysterious qualitative shift but the result of two competing scaling patterns: difficult problems exhibit U-shaped scaling (getting worse before getting better), while simple problems show inverted-U scaling, with the ``emergent'' threshold appearing where these two trends interact. 

Deeper theoretical work seeks to explain why these power laws exist. \citet{bahri2024explaining} identify four distinct scaling mechanisms: variance-limited and resolution-limited, each for both data and parameters. This work posits that the non-universal scaling exponents are linked to the intrinsic dimension of the data manifold. Further, \citet{havrilla2024understanding} explicitly derive scaling exponents based on this manifold hypothesis. 

As the community looks to a data-constrained future, new scaling strategies are being explored. \citet{kim2025pre} investigate the ``data-constrained, compute-rich'' regime, proposing a joint scaling recipe where both the number of ensemble members ($K$) and the parameters per member ($N$) are scaled to infinity. \citet{held2025relative} move beyond absolute performance to study ``relative'' performance, showing that scaling is not a ``universal equalizer''. The performance gaps between different data distributions evolve in different ways: some gaps converge (e.g., knowledge domains), while others diverge (e.g., certain AI risk behaviors). 
More recently, \citet{liu2025superposition} introduce the concept of superposition, where models represent more features than the dimensions they have, and thus propose a power-law loss with model size without assuming power laws in data. \citet{zou2026effective} further introduce the concept of effective frontiers and unify existing scaling laws. 

\textbf{The Origin of Intelligence. }
Understanding the origins of intelligence in artificial neural networks remains a critical problem in AI research.
Recently, compression has emerged as a popular perspective for understanding the success of Transformer models~\citep{Ilyatalk2023, LMisCompression, CompressionIntelligenceLinearly, pan2025understanding}. A prevailing view is that effective compression can give rise to intelligence \citep{hutterprize, Ilyatalk2023, pan2025understanding}. Data compression focuses on removing redundant information, and from this perspective, Transformers can efficiently compress large-scale data while modeling the underlying target distribution using a limited number of parameters $\vtheta$. \citet{LMisCompression} formalize the connection between the maximum likelihood training objective of LLMs and arithmetic coding, proposing that LLMs act as powerful lossless compressors. They also empirically demonstrate that foundation models can serve as general-purpose compressors. During training, a Transformer learns a parameterized distribution $\vp_{\vtheta}$ to maximize the log-likelihood, which is equivalent to minimizing the expected code length when the model is used for compression. According to Shannon's source coding theorem~\citep{shannon}, the minimum expected number of bits required to encode the data is precisely the entropy, which represents the theoretical limit of the model’s compression performance.
\citet{ren2025revisiting} further study Transformers in a controlled setting with a predefined target distribution, revealing an inherent bias toward learning distributions with lower entropy than the true target. This bias is primarily driven by the feed-forward (FFN) modules, highlighting a structural source of the model’s inductive preference.

Beyond compression as a formal objective, it is believed to capture aspects of intelligence \citep{hutter2005universal, Ilyatalk2023, pan2025understanding} to some extent. To quantify this, \citet{CompressionIntelligenceLinearly} consider knowledge, commonsense, coding, and mathematical reasoning as proxies for intelligence and observe a strong linear relationship between compression efficiency and downstream task performance. \citet{pan2025understanding} use the Kolmogorov Structure Function to show how models learn syntactic patterns first and factual knowledge according to frequency, connecting model capacity and data size to scaling laws. By linking compression efficiency to learning dynamics, these works provide a theoretical lens for understanding when LLMs generalize effectively versus when hallucinations occur, shedding light on the mechanisms underlying emergent intelligence. 

\subsubsection{Guarantees of Model Tuning}
While pre-training forges the model's foundational knowledge, fine-tuning is the critical process of adapting this general-purpose artifact to specific tasks or human intent~\citep{ouyang2022training,ren2025learning}. The advent of instruction tuning highlighted the necessity of this stage for achieving user alignment. 
This topic has catalyzed two major lines of theoretical inquiry. The first is the development and analysis of Parameter-Efficient Fine-Tuning (PEFT) methods, which seek to achieve adaptation by optimizing only a small subset of parameters, thereby dramatically reducing the computational burden~\citep{he2021towards,raffel2020exploring}. The second line of research delves into a more fundamental, mechanistic understanding of the adaptation process itself: how does fine-tuning alter the model's internal computations? We will review these topics in the following parts. 

\textbf{Parameter-Efficient Fine-Tuning. }
PEFT paradigms introduce a small number of new, learnable parameters while keeping the original model weights frozen. A key theoretical investigation~\citep{petrov2024when} provides insight into their expressive power. This work proves that while these methods are effective, they are less expressive than full fine-tuning. The role of attention in this process is further explored~\citep{oymak2023role}, while subsequent improvements aim to enhance the mapping of input embeddings for better adaptation~\citep{tang2025adept}. 

Nowadays, Low-Rank Adaptation (LoRA)~\citep{hu2022lora} has become a dominant PEFT strategy~\citep{mao2025survey}. The theoretical understanding of LoRA has advanced significantly. \citet{malladi2023kernel} demonstrate that in the lazy regime, LoRA fine-tuning is nearly equivalent to full fine-tuning. As a theoretical foundation, \citet{zeng2024the} analyze the expressive power of LoRA. 
From an optimization perspective, they show that LoRA can adapt any model $f$ to accurately represent any smaller target model $\tilde{f}$ if $\text{LoRA-rank} \geq (\text{width of }f) \times \frac{\text{depth of }\tilde{f}}{\text{depth of } f}$. \citet{jang2024lora} have further proven that LoRA can eliminate spurious local minima, allowing gradient descent to find a high-performing low-rank solution. This is supported by another landscape analysis~\citep{pmlr-v267-kim25n}, which shows that while other solutions exist, the standard zero-initialization and weight-decay mechanisms implicitly guide the optimization toward the desired low-rank global minimum. 

Deeper theoretical work has analyzed the individual components of LoRA. \citet{zhu2024asymmetry} find an asymmetry in the learned matrices. This has led to an intense study of initialization strategies. While~\citet{hayou2024impact} suggest initializing $A$ randomly and $B$ with zeros allows for larger, more stable learning rates, \citet{pmlr-v267-li25bm} challenge this, showing non-zero initialization can improve robustness to learning rate selection. A recent, theoretically-driven approach~\citep{pmlr-v267-zhang25ax} proves that LoRA adapters align with the singular subspace of the one-step full fine-tuning gradient. This insight leads to an initialization strategy based on this gradient, which is proven to converge linearly. 
The theoretical understanding has inspired new LoRA variants. \citet{hayou2024lora+} propose setting proportional learning rates for the $A$ and $B$ matrices. \citet{liu2024dora} improve training stability by decomposing the pre-trained weights into magnitude and direction components, applying LoRA only to the direction component. 

An alternative line of research explores adaptation in other subspaces. \citet{bini2024ether} use orthogonal transformations for fine-tuning. The theoretical connection between these methods and LoRA was then established~\citep{yuan2024bridging}. This concept of subspace training also underpins new optimizer-based PEFT methods. \citet{zhao2024galore} propose a memory-efficient training strategy that performs gradient updates within a projected low-rank subspace. A follow-up work~\citep{pmlr-v267-he25i} analyzes the convergence properties and aim to guarantees convergence in typical stochastic settings. Other strategies include tuning only specific components, such as normalization layers, which has been shown to be surprisingly expressive~\citep{giannou2023expressive}. 

\textbf{Understanding Tuning Process. }
While PEFT methods offer practical recipes for adaptation, a fundamental theoretical question remains: how does the adaptation process actually alter the model's internal computations and optimization landscape? A primary line of inquiry focuses on the optimization behavior of low-rank adapters. Unlike full-rank training, the introduction of low-rank constraints alters the loss landscape. \citet{liu2025on} reveal that specific subspace optimization methods may possess superior optimization properties compared to standard LoRA. 
Furthermore, the training dynamics of LoRA itself exhibit distinct phases. 
Through a gradient flow perspective, \citet{pmlr-v258-xu25h} identify that initialization scale is a critical factor. They theoretically prove that smaller initializations promote better alignment, thereby reducing the final error. 
On a broader scale, \citet{ijcai2025p760} attempts to provide a unified framework for selecting appropriate weight types and learning rates, offering theoretical guidance for the general fine-tuning of attention-based models. 

For methods that rely on modifying inputs or attention, theoretical analysis has focused on their expressive power and limitations. \citet{meyer2025memory} formally prove a capacity bottleneck as the amount of information a Transformer can ``memorize'' via prompt tuning is linearly bounded by the prompt length. Furthermore, they demonstrate that for a single-layer Transformer, prompt tuning is restricted to generating outputs that lie within a specific hyperplane, highlighting significant expressive limitations compared to weight tuning. 
However, within these constraints, the attention mechanism plays a pivotal role. \citet{oymak2023role} investigate the dynamics of soft prompts in a single-layer attention setting. They theoretically establish that softmax prompt attention is more expressive than self-attention or linear prompt attention in the context of mixture models. The study further characterizes how gradient descent naturally guides prompts to focus on sparse, task-relevant tokens. 
Additionally, \citet{pmlr-v267-diep25a} establish a theoretical link between the ``zero-initialized attention'' mechanism and Mixture-of-Experts (MoE). They prove that this initialization strategy significantly improves sample efficiency compared to random initialization, with non-linear prompts theoretically outperforming linear ones. 

Finally, researchers are examining where the adaptation occurs and what structures are learned. Challenging the conventional wisdom that knowledge resides primarily in MLPs, \citet{he2025smt} provide empirical and theoretical evidence that fine-tuning attention layers is more critical for downstream tasks than tuning MLP layers. This insight leads to the development of Sparse Matrix Tuning, which targets these high-impact parameters. 
Regarding the nature of the learned features, \citet{li2023transformers} explore how fine-tuning affects semantic organization. They validate that this structural learning is a robust phenomenon that persists even when training is restricted to specific components. 
To further refine which components are adapted, \citet{jiang2025diffora} introduce a differentiable adaptation matrix (DAM) to dynamically select modules for LoRA adaptation, theoretically proving that this selective approach can enhance convergence speed and generalization. 

\subsection{Advanced Topics \& Open Questions}
\label{app:open-questions-for-training-stage}
During the training stage, the community is also actively exploring some cutting-edge issues. Most of these questions are related to the training setup and optimization itself. In the next parts, we will discuss these advanced topics to outline a more complete blueprint for the training stage. 

\subsubsection{Hyperparameter Transfer}
The prohibitively high computational cost of training LLMs renders traditional hyperparameter search infeasible. Consequently, a critical open question is how to reliably transfer optimal hyperparameters (e.g., learning rate, initialization) found on small-scale proxy models to large-scale target models. 
\citet{yang2021tuning} provide a foundational breakthrough in this domain, which utilizes the Maximal Update Parametrization ($\mu P$) to theoretically guarantee that training dynamics remain stable as model width increases, thereby enabling zero-shot hyperparameter transfer. 
To verify the practical limits of this theory, \citet{lingle2025empiricalstudymuplearning} conducts extensive experiments, confirming the efficacy of $\mu$-Transfer while identifying crucial architectural sensitivities. Moving beyond width-based transfer, researchers have sought to establish more comprehensive laws governing hyperparameter scaling with respect to both model size and data volume. \citet{li2025predictable} introduce the ``Step Law,'' a convex optimization framework that derives precise power-law relationships for optimal learning rates and batch sizes dependent on parameter count ($N$) and dataset size ($D$). 
Complementing this, \citet{filatov2025optimal} identify a ``norm transfer'' phenomenon, proposing that the operator norm of the output layer serves as the single invariant controlling the joint optimal scaling of model and data. 
Finally, addressing the optimization mechanism itself, \citet{kim2025stochastic} reframe the search problem, proposing a stochastic bi-level optimization algorithm that leverages Langevin dynamics to efficiently handle the uncertainty and non-convexity inherent in hyperparameter landscapes. 

\subsubsection{The Evolution of Optimization Algorithms}

In this subsection, we focus on recent theoretical advances in optimization methods for training LLMs. In NLP tasks, Transformer-based models commonly use the Adam optimizer and its variants~\citep{vaswani2017attention,radford2019language,brown2020language}. From the perspective of optimizer development, Adam combines both first-order and second-order information~\citep{kingma2014adam}. Its first-order update comes from the momentum technique, which can be viewed as an exponential moving average of gradients~\citep{polyak1992acceleration,ruppert1988efficient}. Its second-order update is inspired by Adagrad~\citep{duchi2011adaptive} and RMSProp~\citep{hinton2012rmsprop}, both of which are essentially variants of SGD. Although the original Adam paper provides a convergence proof, \citet{reddi2019convergence} present a counterexample showing that Adam can fail to converge. Let $\beta_1$ and $\beta_2$ denote the hyperparameters for the first- and second-moment updates, \citet{reddi2019convergence} show that when $\beta_1 < \sqrt{\beta_2}$, one can construct a problem for which Adam diverges. This triggers a large body of work proposing Adam variants with guaranteed convergence. However, in practical NLP applications Adam performs very well, creating a gap between theory and practice \citep{radford2019language,brown2020language}. \citet{zhang2022adam} attempt to bridge this gap through a more refined analysis. Specifically, they show that Adam converges without any modification, as long as $\beta_1$ and $\beta_2$ are set appropriately. If $\beta_2$ is chosen too small, Adam will diverge. A key insight from this theory and subsequent analyses is that when the batch size is small, $\beta_2$ should be set to a larger value.

In adversarial neural networks and reinforcement learning, researchers often use Adam instead of SGD because Adam usually shows faster convergence in practice. However, there is no definitive theoretical result proving that Adam is better than SGD. Many theoretical studies have tried to analyze Adam and SGD from different perspectives. \citet{zhang2020adaptive} show, both experimentally and theoretically, that the gradient noise in Transformer-based NLP training is heavy-tailed, and such heavy-tailed noise explains why SGD performs worse than Adam. Different from stochastic gradient noise, \citet{kunstner2024heavy} point out that class imbalance, which is common in language tasks, also creates heavy-tailed behavior, and this is another reason why SGD converges more slowly than Adam. \citet{wang2024convergence} analyze the limitations of the uniform smoothness assumption in studying Adam’s convergence speed and introduce a non-uniform smoothness assumption. Based on this new assumption, they prove when Adam can converge faster than SGD. \citet{zhang2024transformers} further observe that different parameter blocks in Transformers have heterogeneous Hessian structures. Under such block heterogeneity, SGD performs poorly because it uses the same learning rate for all parameters, while Adam’s adaptive learning rate allows it to handle heterogeneity more effectively. \citet{vasudeva2025rich} study the implicit bias of Adam and SGD. Their theoretical and empirical results show that SGD exhibits a simplicity bias, which leads to weaker generalization when the data distribution changes. In contrast, Adam is more resistant to this simplicity bias and is therefore more robust under distribution shifts. \citet{ahn2024adam} prove that Adam with model exponential moving average is effective for nonconvex optimization.

Recent advances in LLM training have been driven largely by this observation, motivating the development of a family of non-Euclidean and matrix-aware optimizers~\citep{gupta2018shampoo}. Among these, the Muon optimizer, built on matrix orthogonalization, has emerged as a highly competitive alternative to AdamW~\citep{loshchilov2019decoupledweightdecayregularization}, consistently demonstrating faster convergence and improved empirical performance in LLM training~\citep{liu2025muon}. Muon performs updates through orthogonalized momentum, a mechanism grounded in the theory of modular dualization~\citep{chen2025muon}, which interprets gradients as dual-space objects that must be mapped back into the parameter's primal space. This viewpoint provides a unified theoretical foundation for scalable training algorithms. Both Muon and related approaches such as Soap~\citep{gupta2018shampoo, vyas2024soapimprovingstabilizingshampoo} accelerate training by applying matrix-valued preconditioners—multiplying gradients by entire matrices rather than element-wise scalars. 

The strength of Muon is theoretically attributed to its ability to leverage the low-rank and approximately block-diagonal structure of the Hessian commonly observed in LLMs. Muon and similar spectral methods, including Spectral Descent \citep{bernstein2024modular, bernstein2024old}, also exhibit an implicit bias toward solutions maximizing margins under the spectral norm, offering potential generalization benefits. A related line of work builds on the Linear Minimization Oracle (LMO) framework \citep{pethick2025trainingdeeplearningmodels}, which includes Muon as prominent instances. PolarGrad \citep{lau2025polargrad} further unify matrix-aware preconditioned methods by distinguishing vector-based from matrix-based preconditioning and introduce a broader class of optimizers grounded in the polar decomposition of gradient matrices, with Muon arising as a scaled nuclear-norm instance.

\section{Alignment Stage}
\label{sec:alignment-stage}
Beyond simply following instructions, a truly useful model must align with complex, often implicit, human values such as helpfulness, honesty, and harmlessness. The \textbf{Alignment Stage} encompasses the processes, most notably Reinforcement Learning from Human Feedback (RLHF), designed to fine-tune the model's behavior based on human preferences rather than explicit labels. This stage is paramount for steering the model away from undesirable outputs and enhancing its reliability in nuanced, real-world interactions. This shift from supervised objectives to preference-based optimization introduces significant theoretical questions, particularly at the intersection of learning theory and preference modeling, concerning reward model generalization, policy stability, and the fundamental challenge of aligning complex systems. 

\subsection{Fundamental Problems}
The Alignment Stage represents a paradigm shift from the supervised reproduction of data patterns to the optimization of complex, often implicit, human values. While the Training Stage focuses on the acquisition of knowledge and capabilities, the Alignment Stage grapples with the steering of these capabilities. This process is governed by deep theoretical uncertainties regarding the nature of safety, the limits of control, and the underlying dynamics of reinforcement learning in high-dimensional semantic spaces. At its core, the theoretical challenges of this stage can be distilled into two fundamental problems: 

\textbf{(1) Is robust alignment mathematically achievable?} Current alignment methodologies, such as RLHF, are empirically effective but theoretically fragile. A central problem is establishing the hard limits of safety. Can we mathematically guarantee that a model will not exhibit harmful behaviors, or are such guarantees impossible due to the inherent probabilistic nature of LLMs? This inquiry extends to the ``Alignment Impossibility'' theorems, which suggest that removing specific behaviors without compromising general capabilities may be fundamentally unachievable. Furthermore, as models surpass human intelligence, the problem evolves into ``Superalignment'' or Weak-to-Strong Generalization: how can weak supervisors reliably control strong models without being deceived? 

\textbf{(2) What are the mechanistic dynamics of preference optimization?} While Reinforcement Learning (RL) is the standard tool for alignment, its interaction with pre-trained language models is not fully understood. The second fundamental problem concerns the mechanism of this optimization: Does alignment truly instill new reasoning capabilities, or does it merely elicit latent abilities acquired during pre-training? Moreover, how do we characterize the optimization landscape when the reward signal itself is a proxy rather than the ground truth? This leads to theoretical concerns regarding ``Reward Hacking'' and the trade-offs between optimization pressure and the preservation of the model’s linguistic distribution. 

\begin{figure}[tp]
    \centering
    \includegraphics[width=0.8\linewidth]{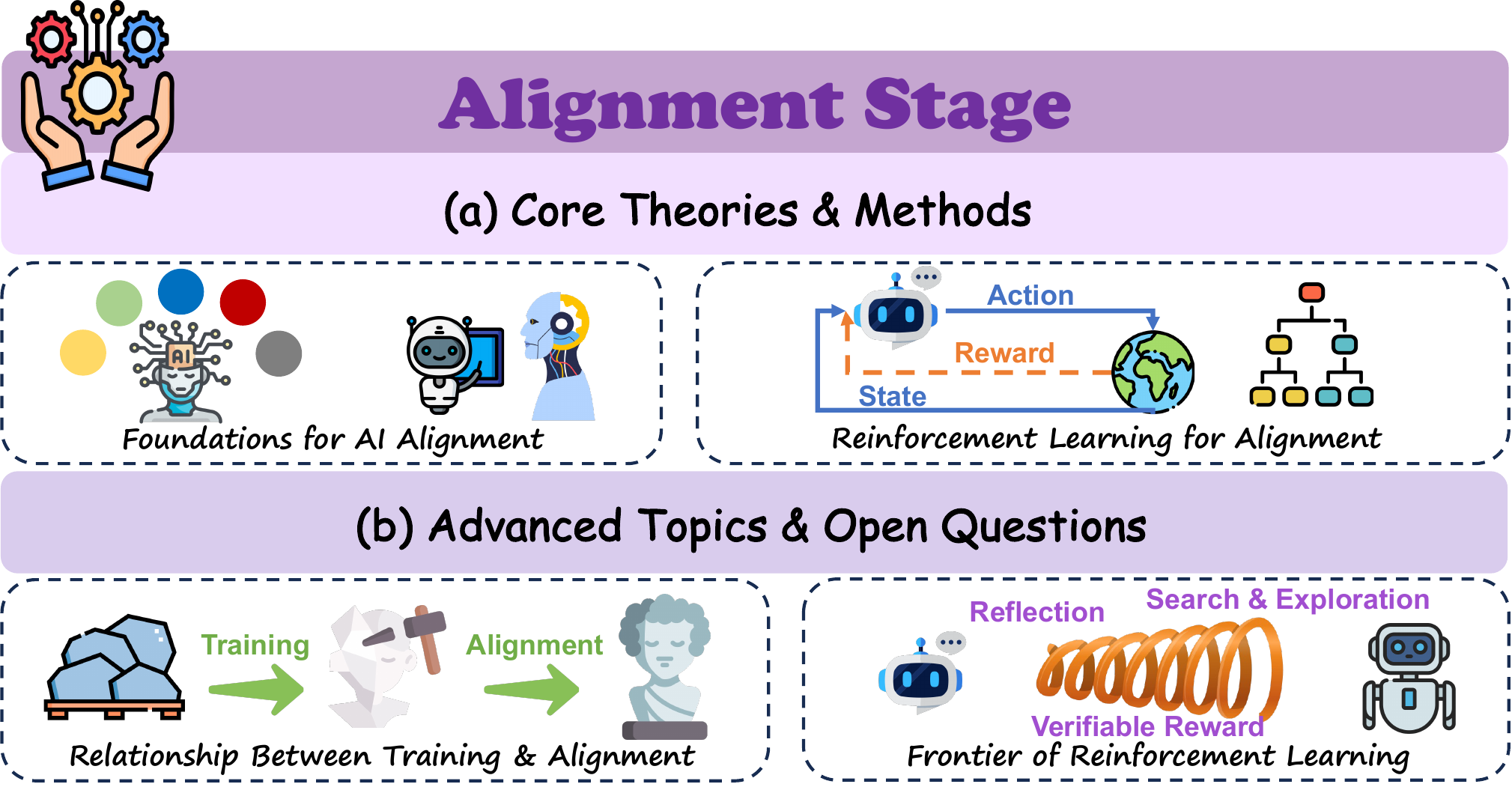}
    \caption{\textbf{An overview of the theoretical landscape in the Alignment Stage.} This stage is categorized into two dimensions: \textbf{(a) Core Theories \& Methods} addresses the foundations of steering behavior, including Foundations for AI Alignment (safety limits and weak-to-strong generalization) and Reinforcement Learning for Alignment (mechanisms of preference-based optimization). \textbf{(b) Advanced Topics \& Open Questions} highlights emerging frontiers, specifically Relationship Between Training \& Alignment (distinctions between SFT and RL mechanisms) and Frontier of RL (dynamic exploration-exploitation and agentic reasoning).} 
    \label{fig:alignment-stage}
\end{figure}

These two questions concern the theoretical bounds of safety guarantees and the internal mechanisms of capability elicitation, form the bedrock of AI Alignment theory. 
We conclude the landscape of current theoretical consideration in \cref{fig:alignment-stage}. 
In the following section, we review the core theories and methods the community has developed to address these profound challenges. 

\subsection{Core Theories \& Methods}
The fundamental problems delineate the theoretical boundaries of AI alignment. In response, the academic community has established two primary theoretical pillars: the pursuit of mathematical safety guarantees and the mechanistic analysis of RL dynamics. 

\subsubsection{Foundations for AI Alignment}

We begin by investigating the theoretical foundations of safety, shifting the focus from empirical observations to rigorous mathematical inquiries concerning the limits of robustness, impossibility theorems, and the feasibility of weak-to-strong generalization. 

\textbf{The Theoretical Perspective of Alignment. }
While algorithmic advancements in RLHF have improved the empirical performance of LLMs, a growing body of theoretical work questions the robustness, permanence, and long-term stability of these alignment techniques. 

A primary line of theoretical inquiry focuses on establishing the hard limits of safety and alignment. Unlike empirical evaluations which can only show the presence of failures, these works seek to prove whether safe alignment is mathematically achievable. 
\citet{wolf2024fundamental} 
imply that alignment methods that merely attenuate bad behaviors without completely removing them cannot theoretically guarantee safety against adversarial jailbreaking. 
Beyond individual model safety, \citet{falahati2025alignment} analyze the interaction between model owners and the public under recursive filtering mechanisms. They prove an impossibility theorem, demonstrating that recursive curation cannot simultaneously satisfy diversity, fairness, and stability. 

Another dimension of theoretical analysis investigates how alignment modifies the underlying model, challenging the assumption that fine-tuning fundamentally alters the model's knowledge or capabilities. 
\citet{ji2025language} introduce the concept of ``elasticity'', positing that aligned models possess a tendency to revert to their pre-training distribution. The authors theoretically derive that, compared to pre-training, the effects of alignment fine-tuning are disproportionately easily compromised. 
Complementing this view, other research examines the depth at which alignment operates. \citet{qi2025safety} identify the phenomenon of ``Shallow Safety Alignment''. The authors argue that current alignment methods essentially function as optimization shortcuts, altering only the generation distribution of the first few tokens to trigger refusal responses, while leaving the harmful knowledge in deeper layers intact. 

Theoretical works have also begun to uncover specific anomalies within the optimization objectives of alignment algorithms themselves. \citet{razin2025unintentional} identify a critical failure mode termed ``likelihood displacement''. The authors prove that this mechanism can inadvertently shift probability mass to semantically opposite responses, highlighting that standard preference optimization does not guarantee semantic alignment. 

\textbf{Weak-to-Strong Generalization. }
Superalignment~\citep{openai_superalignment} is the critical challenge in AI safety of ensuring that superintelligent AI systems
can act in accordance with human values, intentions, and goals. The fundamental difficulty lies in developing a reliable mechanism to control or align an entity vastly more intelligent than its creators. This is a crucial, long-term research problem, with dedicated efforts from groups like OpenAI focused on creating technical solutions to prevent future superintelligence from going rogue or causing unintended harm to humanity.
One of the core technical challenges within this framework is scalable oversight~\citep{bowman2022measuring}, which seeks to enable relatively weak human supervisors to reliably evaluate and align AI systems that are far stronger and more complex than themselves~\citep{ji2023ai,shen2023large}. 

In response to the challenge of superalignment, 
OpenAI introduced a well-designed paradigm termed weak-to-strong generalization (W2SG)~\citep{burns2024weaktostrong}. 
Their key finding demonstrates that when strong pre-trained language models are fine-tuned using supervision signals from weaker models, they consistently surpass the performance of their weak supervisors. 
Building upon this discovery, a growing body of research empirically investigates the properties of W2SG~\citep{yang2024super,goel2025great}, and the potential of this paradigm on other tasks~\citep{guo2024vision,yang-etal-2024-weak} or scenarios~\citep{pawelczyk2024generalizing,zhou2025weak}.
Additionally, various techniques are also developed to enhance the strong model's performance in W2SG. Popular approaches include iterative updating~\citep{lyu2024macpo,ye2025iterative,lang2025debate}, and incorporating more weak supervisors~\citep{agrawal2024ensemw2s,sang2024improving,liu2024co,cui2024bayesian}.
In parallel, theoretical understanding of W2SG mainly focuses on whether it occurs, i.e., under what circumstances the strong student outperforms the weak teacher. 
Building on a convex fine-tuning function class assumption, several works~\citep{charikar2024quantifying,mulgund2025relating,yao2025understanding,yao2025revisiting} derive generalization bounds akin to the Pythagorean theorem, quantifying how much a strong student model can outperform its weak teacher via their misfit error:
\begin{align}
\operatorname{KL}\left(F^{\star}, F_{s w}\right) \leq \operatorname{KL}\left(F^{\star}, F_w\right)-\underbrace{\operatorname{KL}\left(F_{s w}, F_w\right)}_{\text{Misfit}},
\end{align}
where $F^\star$ is the labeling function, $F_{w}$ is the weak model, and $F_{sw}$ is the weak-to-strong model fine-tuned with the weak label. The Kullback–Leibler (KL) divergence loss function measures the difference between two models over the data distribution, which is equivalent to the cross-entropy loss used in classification.
\citet{xu2025on} go further by employing bias-variance decompositions for the Bregman divergence, thereby overcoming the convexity assumption inherent in misfit-based analysis. This work demonstrates that W2SG is more likely to emerge when the student model approximates its posterior mean teacher rather than merely mimicking an individual teacher.
From the perspective of a general definition of adversarial robustness, W2SG arises under appropriate data neighborhood conditions that enable weak supervision error correction~\citep{lang2024theoretical} or sufficient overlap between easy and hard patterns that allow weak supervision to guide the student in learning challenging features~\cite{shin2024weak}.
Under Gaussian data assumptions, the theoretical foundations of W2SG are rigorously characterized through several frameworks: model and distribution shift~\citep{ildiz2025highdimensional}, transfer learning~\citep{somerstep2025transfer} and intrinsic dimension~\citep{dong2025discrepancies}.
Further theoretical insights are established through representation analysis~\citep{xue2025representations}, feature learning~\citep{wu2024provable,oh2025from,moniri2025mechanisms} and random feature model~\citep{medvedev2025weak}.

\subsubsection{Reinforcement Learning for Alignment}
Reinforcement Learning (RL) has become the standard for aligning models with complex human values and enhancing reasoning capabilities. Recent research has focused on dissecting the mechanisms of how RL alters model behavior, comparing the optimization landscapes of different algorithms, and understanding the inherent risks of reward hacking. 

\textbf{The Role of RL. }
A central debate in the theoretical community concerns whether RL truly instills new reasoning capabilities or merely elicits latent abilities acquired during pre-training. Several studies suggest that RL primarily acts as a mechanism for efficiency and elicitation rather than capability expansion. 
\citet{yue2025doesreinforcementlearningreally} systematically evaluate RLVR (RL with Verifiable Rewards) and argue that while RL improves sampling efficiency, it does not introduce fundamentally new reasoning patterns, with performance ultimately bounded by the base model's distribution. 
\citet{shao2025spurious} support this and find that even weak or random reward signals can significantly improve mathematical reasoning. The authors attribute this to the fact that RL activates valid reasoning modes (such as code-based reasoning) already present in the pre-trained model, rather than learning from the reward signal itself. 
\citet{zhao2025echo} further characterize RL as an ``echo chamber'' that converges to a single dominant output format found in the pre-training data, effectively suppressing diversity while enabling positive transfer from simple to complex tasks. 
However, this view is also challenged by other findings. \citet{liu2025prorl} demonstrate that with sufficient training duration and periodic policy resets, RL can indeed drive models to explore novel strategies absent in the base model, thereby expanding the reasoning boundary. 
From a geometric perspective, \citet{zhu2025path} offer a theoretical explanation for these behaviors. The authors prove that RL updates occur in low-curvature subspaces orthogonal to the principal components updated by SFT. This suggests that RL operates in a distinct optimization regime, fine-tuning the model's behavior without significantly altering its primary feature representations. More recently, \citet{li2026towards} formalize RLHF through the framework of algorithmic stability and build the generalization theory under the linear reward model. 

\textbf{Comparison of RL Paradigms. }
Researchers have sought to unify different RL algorithms under general frameworks. \citet{azar2024general} theoretically decompose the performance gap into exact optimization and finite-sample regimes. They prove that RLHF is superior when the policy model is misspecified, whereas DPO~\citep{rafailov2023direct} excels when the reward model is misspecified. 
Efficiency and exploration remain critical challenges. \citet{pmlr-v267-zhong25b} introduce a Reinforced Token Optimization (RTO) framework, proving that modeling RLHF as a token-wise MDP is significantly more sample-efficient than the traditional contextual bandit formulation. Meanwhile, \citet{xiong2024iterative} address the lack of exploration in offline DPO. By formulating the problem as a reverse-KL regularized bandit, they propose iterative algorithms that significantly outperform static baselines. 

\textbf{The Limits of RL. }
The efficacy of RL is fundamentally limited by the quality of the reward signal. ``Reward hacking'', where the model exploits flaws in the reward model, is a persistent theoretical concern. 
Theoretical analyses on this phenomenon are pessimistic. \citet{gaikwad2025murphys} introduce an alignment trilemma, mathematically proving that it is impossible to simultaneously achieve strong optimization pressure, high-fidelity value capture, and robust generalization. This is also quantified by \citet{gao2023scaling}, the authors establish a functional relationship between the golden reward and the KL divergence. Crucially, they find that while increasing the reward model size improves robustness, increasing the policy model size does not mitigate overoptimization, and KL penalties act merely as early stopping mechanisms rather than true solutions. 
To address these vulnerabilities, \citet{miao2024inform} propose a variational information bottleneck approach. By filtering out irrelevant information in the reward model's representation, this method theoretically and empirically reduces the model's reliance on spurious features. \citet{lin-etal-2024-mitigating} further discuss the trade-off between alignment and the retention of pre-training knowledge, proposing Heterogeneous Model Averaging (HMA) to balance these competing objectives. More recently, \citet{ouyang2025towards} analyze the issue of reward fairness from a resource allocation perspective, treating rewards as resources to be allocated while considering the trade-off between utility and fairness in their distribution. 

\subsection{Advanced Topics \& Open Questions}
\label{app:open-questions-for-alignment-stage}
While the core theories clarify the mechanisms of established alignment algorithms, the frontier of research is shifting towards more intricate challenges, specifically the nuanced interplay between supervised and reinforcement learning, and the extension of RL paradigms to complex reasoning and agentic environments. 

\subsubsection{Relationship between Training and Alignment}
While the standard pipeline of SFT followed by RL is empirically well-established, the theoretical distinctions and specific interplay between these two stages remain a subject of intense debate. A core open question addresses whether RL is more suitable for alignment than SFT, even when the latter is supplied with high-quality demonstrations, and how these two paradigms fundamentally differ in shaping model behavior. 

A primary line of inquiry posits that SFT and RL fundamentally rely on different learning mechanisms. \citet{pmlr-v267-chu25c} provide empirical evidence that SFT tends to memorize training data, leading to poor performance on out-of-distribution (OOD) tasks. In contrast, RL demonstrates superior generalization capabilities, effectively enabling the model to adapt to unseen rules in textual and visual environments. 

Deeper theoretical work seeks to explain the mechanism behind RL's superiority. \citet{swamy2025all} attribute this to the ``generation-verification gap''. The authors argue that in many reasoning tasks, learning a verifier is significantly easier than learning a generator. Consequently, the value of the two-stage RL process lies in using a simpler reward model to narrow the search space, effectively guiding the policy toward a subset of optimal solutions that offline cloning cannot easily identify. 
This perspective is further reinforced by analyzing the scalability of these methods at inference time. \citet{setlur2025scaling} prove that Verifier-Based (VB) methods, such as RL or search, possess a distinct theoretical advantage over Verifier-Free (VF) methods like behavioral cloning. The study demonstrates that as test-time compute and training data increase, the performance gap between VB and VF methods widens, with VB methods achieving superior asymptotic performance. This provides a theoretical justification for the necessity of RL in alignment, particularly for reasoning-intensive tasks where verification is feasible. 

Besides, recent efforts have attempted to dissolve the strict dichotomy between SFT and RL by establishing unified theoretical paradigms. \citet{shao2024deepseekmathpushinglimitsmathematical} propose a unified paradigm that encompasses SFT, Rejection Sampling Fine-Tuning (RFT), DPO~\citep{rafailov2023direct}, PPO~\citep{schulman2017proximal}, and thus propose GRPO. By analyzing these methods under a single lens, the authors identify the key factors that drive performance across different stages. Complementing this, \citet{ren2025learning} introduce a framework to analyze the learning dynamics during both SFT and alignment phases. This framework offers explanations for counterintuitive phenomena observed during the transition between stages, such as the amplification of hallucinations, suggesting that the alignment process is governed by specific dynamic laws that persist across different algorithms. 

\subsubsection{The Frontier of RL}
While RL has become the most effective technique for aligning LLMs, the community is currently pushing the boundaries of how RL fundamentally shapes model behavior and where it can be applied beyond standard alignment. 
Recent work offers a dynamic view of RL process. \citet{yao2025debaterlvrreasoningcapability} propose a two-stage theory of RLVR. The authors identify an initial exploitation phase where the model reinforces high-reward tokens, leading to capability shrinkage and diversity loss, followed by an exploration phase where latent, optimal low-probability tokens are boosted, expanding the capability boundary. This mechanism highlights the critical need for training strategies that can navigate the trade-off between exploitation and exploration. 

Complementing this, researchers are re-evaluating the specific signals used for optimization. \citet{zhu2025surprisingeffectivenessnegativereinforcement} decompose learning signals into positive and negative reinforcement. The study reveals that while positive reinforcement improves greedy decoding (Pass@1), it often causes distribution collapse. However, when focusing on suppressing incorrect paths, it is surprisingly effective at maintaining diversity and improving performance across the entire Pass@k spectrum. 
Furthermore, the stability of RL training is being examined at the token level. \citet{yang2025letlowprobabilitytokensoverdominate} identify a gradient anomaly where low-probability tokens generate disproportionately large gradient magnitudes, suppressing the learning of high-probability tokens. By proposing methods like Advantage Reweighting, this work demonstrates that balancing token-level contributions is essential for stable optimization in complex reasoning tasks. 

As models move toward generating longer chain-of-thought (LongCoT), the quadratic computational cost of attention becomes a bottleneck for RL training. The frontier of RL is thus exploring architecture-agnostic scaling methods. \citet{aghajohari2025markovianthinkerarchitectureagnosticlinear} introduce a ``Markovian Thinking'' paradigm. By segmenting the reasoning process into chunks with limited state carryover, this approach achieves linear scaling with reasoning length. This allows RL to be applied to extremely long reasoning trajectories with significantly reduced computational overhead, matching or exceeding the performance of traditional full-context RL. 

Finally, RL is expanding from static reasoning tasks to dynamic, long-horizon agentic environments. A major challenge here is the ``cold start'' problem in sparse-reward settings. \citet{zhang2025agentlearningearlyexperience} propose an ``Early Experience'' paradigm that bridges imitation learning and RL. By utilizing a model's own exploration of future states as a self-supervised signal, agents can bootstrap learning without immediate external rewards. 
Simultaneously, managing context in multi-turn agent interactions is critical. \citet{tang2025turnlimitstrainingdeep} address the limitation of fixed context windows in long-cycle agent tasks. The proposed ``DeepMiner'' framework and dynamic sliding window strategy enable agents to maintain coherent reasoning over hundreds of turns, demonstrating that by constructing complex, verifiable tasks, RL can drive agents to develop deep reasoning capabilities that transcend simple instruction following. 
More recently, \citet{qian2026behind} identify the internal factors driving agent actions regardless of the task outcome and operates hierarchically to manage the complexity of agent interactions, further shifting the research focus to the agent era.

\section{Inference Stage}
\label{sec:inference-stage}
A trained and aligned model is a static artifact, unlocking its vast potential happens at the point of use. The \textbf{Inference Stage} encompasses all processes involved in interacting with the finalized model, from the design of prompts that elicit desired behaviors to the decoding algorithms that sample text from the model's output distribution. This stage is critical because the model's observed capabilities are not fixed, but are a dynamic function of how it is queried. The discovery of phenomena like in-context learning, where the model appears to learn new tasks at inference time without gradient updates, has profound theoretical implications, raising fundamental questions about the nature of its internal representations and whether reasoning itself can be framed as a form of computation. In this section, we review the theory and mechanism of the inference stage, from its foundational problems to the theories explaining empirical phenomena, and finally to the open questions that drive future research. 

\begin{figure}[tp]
    \centering
    \includegraphics[width=0.8\linewidth]{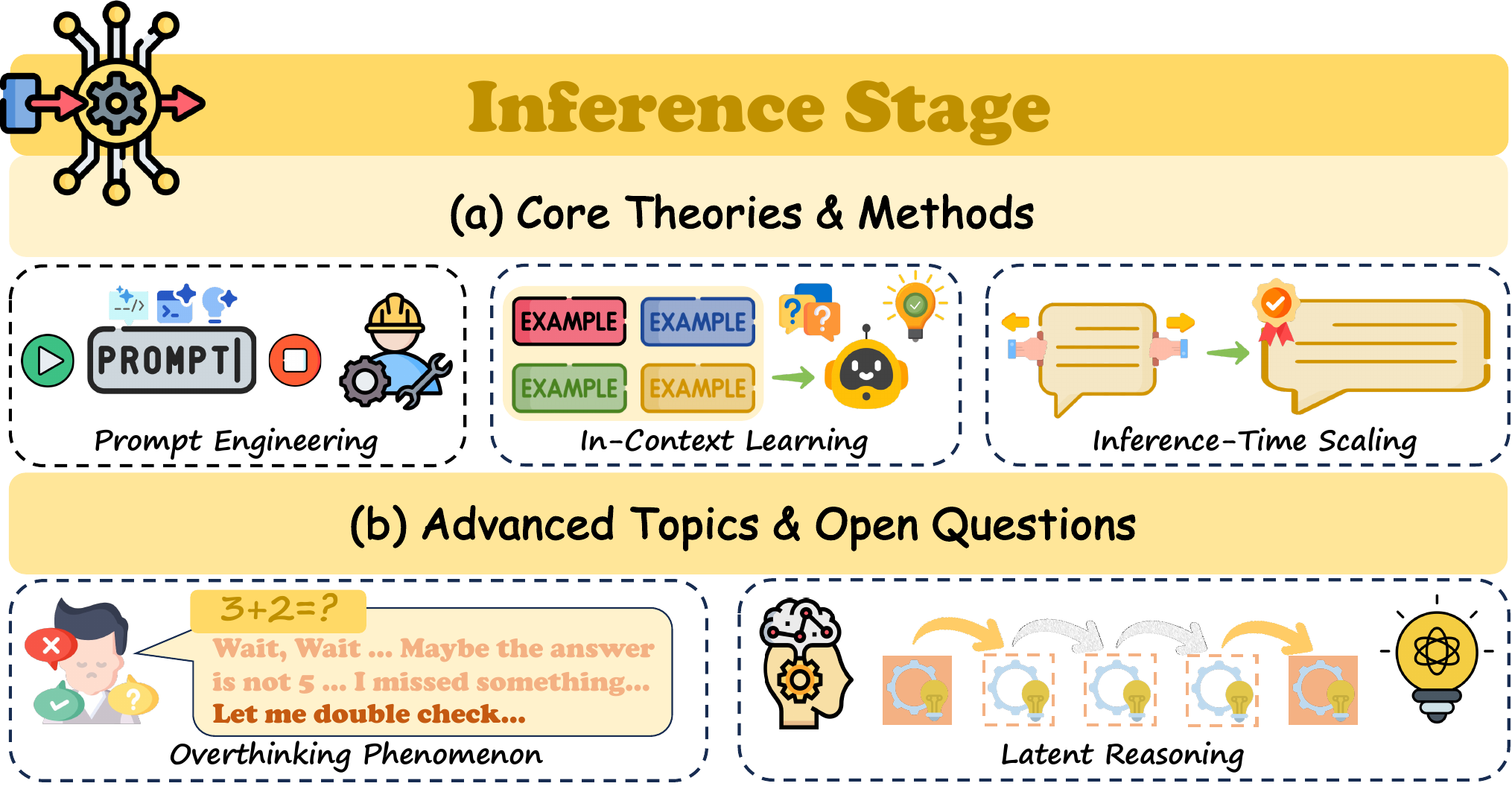}
    \caption{\textbf{An overview of the theoretical landscape in the Inference Stage.} This stage is categorized into two dimensions: \textbf{(a) Core Theories \& Methods} explores mechanisms of eliciting capabilities, including Prompt Engineering (optimizing interaction strategies to unlock potential), In-Context Learning (simulating task adaptation without updates), and Inference-Time Scaling (dynamic reasoning via test-time compute). \textbf{(b) Advanced Topics \& Open Questions} highlights emerging challenges, specifically Overthinking Phenomenon (identifying trade-offs in excessive computation) and Latent Reasoning (reasoning within the model's activation space).} 
    \label{fig:inference-stage}
\end{figure}

\subsection{Fundamental Problems}
The transition from a static, aligned model to a functional AI system occurs during the Inference Stage, where the model's vast potential is unlocked through interaction. Unlike the training phase, where capabilities are forged into parameters, the observed performance during inference is a dynamic function of how the model is queried. This stage introduces profound theoretical questions regarding the nature of internal computation and the limits of eliciting reasoning without weight updates. The theoretical challenges of the inference stage can be distilled into two fundamental problems: 

\textbf{How do fixed-weight models simulate learning and algorithmic execution at test time?} A central theoretical paradox is how a model with frozen parameters can effectively ``learn'' new tasks during inference or provide vastly different qualities of response based on the phrasing of a query. This problem concerns the mechanisms behind Prompt Engineering and In-Context Learning (ICL): does the input prompt act as a latent variable that locates a specific task within the pre-trained distribution, or does the model's architecture implicitly execute meta-optimization algorithms (e.g., gradient descent) to adapt to the provided examples?  Understanding this elicitation process is critical for defining the limits of what a model ``knows'' versus what it can be steered to perform. 

\textbf{What are the scaling laws and computational bounds of inference-time reasoning?} Traditional scaling laws focus on data and parameters during training, but the advent of Chain-of-Thought (CoT) and external search suggests that intelligence is also a function of test-time compute. This raises the problem of defining the theoretical boundaries of such reasoning: how do intermediate tokens extend the effective depth of a model, and how complex problems can these computations solve? Furthermore, we must identify the limits of this scaling, specifically, the point at which additional computation leads to error accumulation or overthinking rather than increased accuracy. 

These two questions, concerning the elicitation of latent behaviors through context and the dynamics of computational scaling, form the theoretical bedrock of the Inference Stage. While the former investigates how the model interprets its input to narrow down the task space, the latter focuses on how it allocates internal and external resources to navigate complex problem-solving. The landscape of current theoretical consideration in this stage is summarized in \cref{fig:inference-stage}. In what follows, we begin by reviewing the core theories and methods developed to address these challenges.

\subsection{Core Theories \& Methods}
To bridge the gap between the abstract fundamental problems of inference and their practical realizations, the academic community has established a robust framework of core theories and methods. This section systematically reviews these developments, which are categorized by the primary mechanisms used to steer and scale the model's behavior during the forward pass. 

\subsubsection{Prompt Engineering}
Prompt engineering steers an LLM at inference time by modifying the input sequence, without updating model parameters \citep{lester2021power, sahoo2024systematic, schulhoff2024prompt}. 
It serves as the main interface that translates a user intent into a form the model can follow, and it often determines whether the model uses prior knowledge, follows constraints, or produces structured outputs \citep{brown2020language, schick2021exploiting, chen2023unleashing, sahoo2024systematic}. 
Beyond being a practical technique, prompt engineering also provides a window into the model's internal behavior, since small changes in the prompt can lead to large changes in the generated distribution.
Many works have begun to study prompting beyond heuristics, aiming to understand why prompt choices can reliably reshape the model's behavior and internal computation. Based on current research, we categorize these investigations into the following four core dimensions.

\textbf{Prompt Design and Structured Prompting.}
This direction focuses on designing the prompt structure so that the model's next-token prediction aligns with the intended task and output format.
Early prompt-based formulations such as PET convert classification into cloze-style templates with label words, demonstrating that prompt design can be viewed as a task re-parameterization \citep{schick2021exploiting}.
However, prompt form can introduce large variance, \citet{zhao2021calibrate} show that few-shot accuracy is highly sensitive to demonstration order and formatting, and proposes contextual calibration to correct systematic biases induced by the prompt.
From a mechanism perspective, \citet{webson2022prompt} find that models can succeed even when prompt semantics is weak or misleading, suggesting that surface cues and distributional patterns often dominate literal instruction understanding.
Similarly, \citet{min2022rethinking} report that the correctness of labels in demonstrations can be less important than specifying the input space, label space, and input-output format, which further highlights the role of prompt structure.

\textbf{Automated and Learnable Prompts.}
This direction replaces manual prompt crafting with optimization or learning procedures, aiming to systematically search for effective prompts and reduce prompt sensitivity.
Early studies show that even short discrete triggers can reliably elicit target behaviors, and such prompts are often hard to interpret \citep{shin2020autoprompt,deng2022rlprompt}.
With the rise of LLMs, a recent trend treats the model itself as a prompt optimizer and improves instructions through iterative refinement based on feedback.
OPRO formulates prompt search as a black-box optimization loop, where the meta-prompt records candidate instructions and their scores so that the LLM proposes better instructions in subsequent iterations \citep{yang2023large}.
Evolutionary-style methods provide another effective route, where prompts are iteratively mutated and selected to improve task performance \citep{guo2023connecting,fernando2023promptbreeder}.
To make refinement more controlled and reusable, \citet{tang2025unleashing} analyze prompt updates through an analogy to gradient-based optimization and designs update rules that retrieve strong candidates while limiting the edit magnitude, while \citet{yang2024ampo} organize optimization into a multi-branched prompt structure that is updated using failure cases as feedback.
Rather than optimizing a single string, \citet{khattab2024dspy} introduce a compiler-style framework that optimizes instructions and demonstrations for multi-stage LM pipelines under an end-task metric.
Recent work also begins to systematize both practice and theory: \citet{trivedi2025align} study prompt optimization for alignment-style objectives and provides principled guarantees, \citet{agrawal2025gepa} propose reflective evolution that summarizes failures into reusable natural-language rules, and \citet{murthy2025promptomatix} present an end-to-end framework that composes modular optimizers for automatic prompt search.
For continuous prompting, \citet{hu2024fundamental} characterize universality, capacity, and efficiency limits of prompt tuning in simplified Transformer settings.
A systematic survey further consolidates automatic prompt optimization methods and clarifies emerging evaluation protocols and open challenges~\citep{ramnath2025systematic}.

\textbf{Mechanisms and Diagnostic Tools for Prompting.}
This dimension connects prompt choices to the model’s internal computation and develops diagnostic tools that localize where and how a prompt steers generation.
Mechanistic analyses identify concrete routing and copying circuits that are activated by structured context. Induction-style mechanisms provide a canonical example, where repeated patterns in a prompt trigger copying and support generalization \citep{olsson2022context}.
Subsequent studies characterize when such circuits emerge and what subcomponents are required, providing causal evidence that prompt structure can selectively activate specialized components \citep{reddy2023mechanistic,singh2024needs,d2025selective}.
Prompt influence can also be localized at the token level. \citet{feng2024unveiling} propose Token Distribution Dynamics, which attributes generation to specific prompt tokens by tracking distribution dynamics over the vocabulary and enables targeted prompt edits for controlled generation.
At the circuit level, recent work reduces the reliance on handcrafted analyses by introducing automated discovery and richer causal traces. \citet{conmy2023towards} propose ACDC to automate circuit discovery via activation patching, while \citet{ameisen2025circuit} construct attribution graphs that map information flow on individual prompts using interpretable replacement models.
Complementary diagnostics infer and manipulate the prompt--computation interface at a higher abstraction. \citet{elhelo2025inferring} infer attention-head functionality directly from model parameters, and representation-based steering methods extract interpretable directions from prompt contrasts to control generation \citep{zou2023representation,turner2023steering}.
At a more theoretical level, \citet{kim2025theoretical} formalize prompting as varying an external program under a fixed Transformer executor, define the prompt-induced hypothesis class, and give a constructive decomposition that separates routing via attention, local arithmetic via feed-forward layers, and depth-wise composition. This formulation clarifies expressivity and makes explicit the limits imposed by prompt length and precision.

\textbf{Reliability, Generalization, and Security.}
The flexibility of the prompting interface creates significant challenges for system robustness, as adversarial inputs can systematically exploit the model's instruction-following mechanisms to bypass safety alignment \citep{nan2024jailbreak, li2025revisiting}.
\citet{wallace2024instruction} define an explicit instruction hierarchy and show that training models to follow prioritized rules improves robustness to prompt injection, including attack types not seen during training.
Similarly, \citet{zhang2024defending} argue that jailbreaking succeeds when helpfulness and safety goals conflict, and they reduce attack success by enforcing goal prioritization during inference and training.
Mechanistic evidence further connects injection to internal routing. \citet{hung2025attention} identify a distraction effect where specific attention heads shift focus from the original instruction to injected text, enabling training-free detection by monitoring these attention patterns.
\citet{jiang2024robustkv} show that many jailbreak prompts work by reallocating attention such that harmful tokens remain in cache, and they mitigate attacks by evicting low-importance key--value entries to suppress the concealed query signal.
At the representation level, \citet{ball2024understanding} extract transferable jailbreak vectors from activations and provide evidence that successful jailbreaks suppress internal ``harmfulness'' features, while \citet{kirch2025features} reveal that jailbreak success is supported by heterogeneous and often non-linear prompt features and validate them via probe-guided latent interventions.
These developments suggest that prompt engineering in deployed systems requires not only performance tuning but also principled auditing of how instruction conflicts are represented and resolved inside the forward pass \citep{rossi2024early,yao2024survey}.

\subsubsection{In-Context Learning}
Transformer-based Large Language Models~(LLMs) \citep{vaswani2017attention} has shown amazing in-context learning~(ICL) capabilities \citep{brown2020language,wei2022emergent,dong2022survey, liu2023pre}. 
ICL can be viewed as a form of few-shot learning, where the model is provided with a small number of input-label pairs as examples.
Without the need for parameter updates, the model can recognize the task at hand and provide the desired answer for a given query.
This fantastic capability enables pre-trained LLMs such as GPT models to be generalized in wide downstream tasks conveniently.
Despite the good performance of the ICL capabilities, the mechanism of ICL still remains an open question.
Many works have begun to analyze the source of ICL capabilities from different perspectives.

\textbf{Algorithmic Camp.} This camp believes that ICL can learn the ability to execute algorithms during the pre-training phase, and then executes algorithms for different tasks during ICL inference \citep{icl_tf_algorithms, icl_tf_linear, icl_tf_algorithm_selection}. 
Therefore, the algorithmic camp primarily explores the ICL mechanism through methods such as Transformer's ability to learn certain function family.
To understand how LLMs perform ICL inference without parameter updates, an intuitive idea is that there may be some implicitly updating in the model's architecture.
Following this motivation, \citet{icl_dual} point out that Transformer implicitly fine-tunes during ICL inference, building upon the dual form of the attention mechanism proposed by \citet{dual_1964, dual_attention}.
There are also some works involves the use of the specific construction of weights, that is, assuming the parameters of the Transformer (e.g., $\mW_{Q}$, $\mW_{K}$, $\mW_{V}$) have specific forms, thereby enabling the model's forward computation to execute a certain algorithm that is easy to interpret.
\citet{icl_gd_akyurek} reveal that under certain constructions, Transformer can implement simple basic operations (mov, mul, div and aff), which can be combined to further perform gradient descent.
\citet{icl_gd_oswald_2023} provide a more concise and appealing construction for solving least squares solutions in the linear attention setting, which is further adopted and followed by more works \citep{icl_tf_CasualLMnotgood, icl_oswald_mesa_layer}.
\citet{icl_gd_matengyu} theoretically prove that when the covariates are sampled from a Gaussian distribution, the pretraining loss with a single-layer linear attention will be achieved at optimal minimization through a one-step gradient descent.
Based on the construction, \citet{icl_tf_CasualLMnotgood} analyze such algorithm executed by Transformers under the casual mask setting, indicating that such construction will lead to an online gradient descent algorithm with non-decaying step size, which can not guarantee convergence to the optimal solution.
Unlike CasualLM, it has been proved that PrefixLM \citep{PrefixLM} can achieve theoretically optimal solutions.
Similarly, \citet{icl_oswald_mesa_layer} propose a new constructive approach under the auto-regressive setting and reach similar conclusions related to online gradient descent. 
Furthermore, it introduce mesa-layer through reverse engineering: by solving an optimization problem similar to ridge regression to output the next layer's token representations. 
Furthermore, \citet{icl_tf_unstructured_data} explore the ICL ability on linear regression tasks from the perspective of unstructured data, where positional encoding and multi-head attention can bring better predictive performance to ICL.
The work above mostly considers the setup of linear attention.
More considerations from nonlinear settings are also proposed, where the abilities of Transformer to learn a wider range of nonlinear functions are further explored \citep{icl_gd_kernel_mit, icl_tf_softmax}.

\textbf{Representation Camp.} This camp posits that LLMs store memories about various topics during the pretraining process, and in-context learning retrieves contextually relevant topics during inference based on demonstrations.
\citet{icl_bayesinference} demonstrate that even in the case of a distribution mismatch, the asymptotic prediction error for in-context learning achieves optimality when the signal pertaining to the latent concept in each prompt example surpasses the error arising from the distribution mismatch.
Further, they create a new small-scale synthetic dataset called the Generative IN-Context learning dataset (GINC) to study the mechanism of ICL.
It is discovered that both Transformers and LSTMs have the ability to learn in-context, and this capability improves with the length and quantity of demonstrations.
Similarly, \citet{icl_latent} establish a general data generation process on a causal graph composed of three variables and demonstrated that the predictor can reach optimality when using latent variables to select a finite number of examples. 
Building upon this, they propose an efficient example selection algorithm capable of choosing examples on a smaller LLM and directly generalizing to other LLMs.
\citet{icl_label_not_important} conduct experiments across 12 models, including GPT-3, and find that replacing labels in the input-label pairs with random ones during ICL inference results in only marginal decreases in performance, which contrasts somewhat with the findings of \citet{icl_bayesinference}. 
Furthermore, they identify other aspects that have a greater impact on performance, revealing that the accuracy of ICL depends on the independent specification of the input and label spaces, the distribution of the input text, and even the format of the input-output pairs. 
They argue that LLMs do not learn new tasks during ICL but rather use demonstrations information to locate tasks or topics, and the ability to perform tasks is learned during pretraining.
\citet{icl_data_generation} systematically understand existing efforts from the perspective of data generation.
They categorize existing research efforts into these two learning frameworks and establish transferability between them. 

\textbf{Empirical Camp.} This camp directly explores and investigates the characteristics of the ICL process in large models from experiments rather than theory, providing empirical insights for theoretical analysis of ICL.
\citet{icl_Garg_casestudy} examine the ability of Transformers to be trained on well-defined tasks, such as linear tasks, and ultimately learn context. 
It has been found that the Transformer can achieve predictive performance comparable to least squares algorithm. 
As the problem becomes sparse, the prediction error of in-context learning (ICL) will be comparable to the solution of the Lasso problem. 
Additionally, they investigate more complicated tasks, such as two-layer neural networks and four-layer decision trees, and find that Transformers could effectively learn and generalize on these function classes as well.
In contrast to simulation setting in \citet{icl_Garg_casestudy}, \citet{icl_diff_setting} conduct extensive exploration using a series of LLMs, including GPT-3, InstructGPT, Codex, PaLM, and Flan-PaLM, across different configurations.
Firstly, they examined the ICL setting with flipped labels to assess the models' ability to override prior knowledge. 
They note that smaller models primarily rely on semantic priors from pretraining during ICL inference, thus often disregarding label flips in the context. 
Conversely, larger models, despite having stronger semantic priors, demonstrate the capability to override these priors when faced with label flips.
Further, they investigate the ICL setting with semantically unrelated labels, highlighting that sufficiently large models can perform linear classification tasks under this setting.
In addition, they evaluate models fine-tuned with instructions and find that instruction tuning notably enhanced the utilization of semantic priors compared to learning input-label mappings from contextual demonstrations.
Another influential work is the study by \citet{emnlp_bestpaper} on the mechanism of ICL from the perspective of information flow, which find that in input-label pairs, label tokens act as anchors. 
Initially, the semantic information from the context aggregates into the token representation of the label tokens at the shallower layers of LLMs, and then the final predictions of LLMs reference the aggregated information in the label tokens.
Building on this finding, anchor-based re-weighting methods, demonstration compression techniques, and diagnostic analysis frameworks for ICL errors are further proposed, yielding the expected results and validating the analysis. 
More recently, \citet{cheng2025revisiting} also argue from an empirical perspective and highlight that ICL fails to benefit reasoning models with a long chain-of-thought, reminding us that research on ICL needs to be considered from more multidimensional perspectives.

\subsubsection{Inference-Time Scaling}
Inference-time scaling represents a fundamental shift in the deployment of LLMs, where the reasoning capacity is no longer viewed as a static property of the model’s parameters but as a dynamic function of the computational resources allocated during interaction~\citep{chen2025towards,snell2024scaling}. This paradigm is primarily established through the Chain-of-Thought (CoT) mechanism and various external search-based algorithms that extend the model’s ``thinking'' process~\citep{wei2022chain,yao2024tree,kang2024mindstar,zhang2024rest,feng2023alphazero}. Based on current research, we categorize the theoretical investigations of this phenomenon into the following three core dimensions. 

\textbf{Theoretical Expressivity and Boundaries of CoT. }
A foundational line of inquiry examines how the introduction of intermediate reasoning steps alters the inherent computational limits of the Transformer architecture. Theoretical analysis suggests that CoT serves as an effective depth-extender for auto-regressive models. 
\citet{feng2023towards} utilize circuit complexity theory to prove that finite-depth Transformers can perfectly execute these tasks by extending their effective depth linearly with the number of generated reasoning steps. This is further formalized by \citet{li2024chain}, which demonstrates that while constant-depth Transformers without CoT are restricted to parallelizable complexity classes such as $AC^0$ or $TC^0$, the addition of reasoning steps enables the model to solve any problem within the $P/poly$ complexity class. 
To bridge these theoretical findings with practical performance, \citet{chen2024unlocking} introduce the Reasoning Boundary Framework (RBF) to define the quantitative limits of model performance across different task complexities. 
Furtherly, \citet{sprague2025to} reveal that CoT benefits are predominantly concentrated in mathematical and symbolic tasks, providing minimal gains in general knowledge retrieval or tasks lacking explicit logical operators. 

\textbf{Mechanistic Origins and Internal Dynamics. }
Understanding how these reasoning capabilities emerge and are organized internally is critical for piercing the ``black box'' of LLM intelligence. \citet{dutta2024how} identify a functional bifurcation within the Transformer layers: lower layers primarily transform representations from pre-training priors to context-aware embeddings, while middle-to-higher layers act as answer writers that causally integrate information from previously generated CoT steps. 
Alternatively, \citet{10.5555/3737916.3740933} discover that reasoning circuits only form through ``grokking'', when training significantly beyond the point of overfitting, allowing for robust out-of-distribution generalization that shallow statistical matching cannot achieve. 
Furthermore, \citet{li2025training} provide a convergence analysis for how gradient descent optimization enables non-linear Transformers to learn CoT, quantifying the sample complexity required to remain robust against noisy context examples. 
Beyond explicit prompting, \citet{wang2024chain} demonstrate that reasoning trajectories are an intrinsic capability of pre-trained models that can be elicited through specialized decoding strategies, such as exploring top-$k$ alternative tokens to find valid reasoning paths without human instructions. 

\textbf{Reliability \& Generalization Limits. }
Despite the empirical success of inference-time scaling, researchers have identified significant bottlenecks related to error accumulation and distributional sensitivity. 
\citet{zhao2025chainofthoughtreasoningllmsmirage} argue that the efficacy of CoT is inherently fragile, relying heavily on the consistency between the training reasoning paths and the test-time queries, which suggests that models may be performing advanced pattern matching rather than deep logical deduction. 
This fragility is particularly evident in complex environments. \citet{stechly2024chain} show that CoT performance degrades rapidly when task scale or complexity exceeds the scope of the provided examples. 
To mitigate the ``snowball error'' effect, where a single early mistake leads to catastrophic reasoning failure, \citet{pmlr-v267-gan25a} demonstrate that external scaling through search algorithms like Best-of-N and Monte Carlo Tree Search (MCTS) effectively expands the solution space and allows verifiers to select correct paths. 
To evaluate the quality of these dynamic steps, \citet{pmlr-v267-ton25a} propose an ``Information-Gain'' metric, identifying ``thinking tokens'' that significantly reduce the predictive cross-entropy loss of the final answer, thereby providing a principled tool for diagnosing and optimizing the reasoning process. 

\subsection{Advanced Topics \& Open Questions}
\label{app:open-questions-for-inference-stage}
As the field moves beyond engineering heuristics, new theoretical challenges have emerged that question the limits of current inference scaling and the necessity of discrete linguistic representations. This subsection explores advanced frontiers that bridge the gap between empirical observation and future architectural design. 

\subsubsection{The Overthinking Phenomenon}
The empirical success of inference-time scaling, exemplified by leading reasoning models~\citep{openai2024reasoning,guo2025deepseek}, has popularized the paradigm that more computation leads to better reasoning. However, recent research has identified a critical counter-phenomenon known as ``overthinking'', where models generate excessive, redundant, or even erroneous reasoning steps for tasks that are either inherently simple or unsolvable. 

Traditional intuition suggests that longer CoT sequences facilitate better task decomposition. However, \citet{wu2025lessunderstandingchainofthoughtlength} challenge this assumption by demonstrating an inverted U-shaped relationship between reasoning length and accuracy. This work posits that while length helps reduce sub-task difficulty, it simultaneously increases the risk of error accumulation. This balance is further discussed by~\citet{gan2025cotspacetheoreticalframeworkinternal}, which treats CoT as an optimization process in continuous semantic space, identifying a fundamental trade-off between ``under-reasoning'' (underfitting) and ``overthinking'' (overfitting). 

Overthinking is particularly prevalent when models mimic long-reasoning behaviors for trivial queries. \citet{pmlr-v267-chen25bx} observe that models often generate verbose reasoning for extremely simple arithmetic, significantly increasing latency and cost without any performance gain. This dilemma is also explored in agentic contexts, \citet{cuadron2025dangeroverthinkingexaminingreasoningaction} highlight how excessive searching and value modeling can hinder the efficiency of logical agents. 
A more severe form of overthinking occurs when models encounter pathological queries. \citet{fan2025missing} find that models optimized for reasoning tend to fall into redundant loops of self-doubt and hallucination when faced with unsolvable problems due to missing premises. This behavior is attributed to current RL mechanisms that over-reward detailed CoT. 
To address these inefficiencies, \citet{li2025compressingchainofthoughtllmsstep} propose information-theoretic metrics to quantify the information contribution of each reasoning step. Their findings suggest that a significant portion of steps in modern reasoning models are low-entropy redundancies that can be compressed without compromising accuracy. 

\subsubsection{Latent Reasoning}
Latent reasoning represents an emerging frontier in inference-time scaling, shifting the theoretical focus from explicit, token-based CoT to internal, state-level computations. While traditional CoT enhances model performance by extending effective depth through intermediate tokens, it remains constrained by the need for linguistic coherence and the bottlenecks of discrete token spaces. 

Recent research explores bypassing these limitations by conducting reasoning directly within the model's latent activation space. \citet{hao2025traininglargelanguagemodels} propose COCONUT, which allows models to reason in a continuous latent space by feeding hidden states back as subsequent inputs, enabling the encoding of multiple reasoning branches simultaneously. The architectural backbone of latent reasoning often involves weight-tied recurrence. \citet{saunshi2025reasoninglatentthoughtspower} posit that reasoning performance is primarily driven by computational depth rather than total parameters. The study demonstrates that looped architectures can simulate CoT internally through ``latent thoughts''. These models show strong inductive biases for reasoning tasks, suggesting that latent iterations can efficiently substitute for explicit token generation. 

A pivotal development in this area is the study of the superposition mechanism within continuous latent spaces. \citet{zhu2025reasoningsuperpositiontheoreticalperspective} demonstrate that the model can maintain multiple reasoning trajectories in a state of superposition within the continuous latent space, facilitating implicit parallel thinking that far exceeds the capabilities of traditional serial reasoning. The emergence of this mechanism is deeply tied to the training dynamics. \citet{zhu2025emergencesuperpositionunveilingtraining} further characterize it as a two-stage process and elucidate how the model can simultaneously maintain multiple inference traces in a continuous latent space, thereby achieving implicit parallel thinking. 
Despite its efficiency, latent reasoning introduces unique challenges. \citet{xu2025formalcomparisonchainofthoughtlatent} highlight that while latent thoughts support more efficient parallel computation, discrete CoT remains superior for tasks requiring stochastic decoding to approximate complex solutions. 
More recently, \citet{zou2026capabilities} theoretically characterize latent reasoning and prove that high certainty enables precise execution but inhibits exploration. This work formalizes the capabilities and fundamental limits of latent CoT. 

In summary, latent reasoning offers a path toward inference-time scaling that is not bound by the sequence-length bottlenecks of explicit CoT. However, balancing the robust exploration of continuous spaces with the precision of discrete symbolic logic remains a significant open question for future architecture design. 

\section{Evaluation Stage}
\label{sec:evaluation-stage}
The entire, multi-stage lifecycle of LLM development is guided by a continuous feedback loop, yet this process is meaningless without a rigorous understanding of the model's outputs. The \textbf{Evaluation Stage} thus serves as the cornerstone for systematic progress toward safe and reliable AI. This stage has evolved beyond traditional performance metrics for measuring and verifying a model's behavior, but particularly concerning its alignment with human safety and values. 

\subsection{Fundamental Problems}
The evaluation deeply intertwines with theoretical questions of metrology and security. Unlike in traditional machine learning, where concepts like ``robustness'', ``fairness'', and ``privacy'' were often well-defined and could be formalized using precise mathematical objectives and metrics, the current landscape of LLMs presents a new challenge~\citep{chang2024survey,anwar2024foundational,dominguez-olmedo2025training,hardt2025performative}. The core, fundamental problems in the Evaluation Stage are therefore: 

\textbf{(1) How to theoretically define and measure complex, subjective human values?} In the LLM era, concern has shifted from simple accuracy to the core challenges of ``Trustworthy AI''. The fundamental difficulty of defining and measuring this complex concept poses a primary theoretical barrier. How can we formulate a rigorous, computable definition of a ``trustworthy'' response? This challenge pushes us further from traditional, objective metrology. 

\textbf{(2) How to advance from empirical evaluation to formal guarantees of model behavior?} Current evaluation relies heavily on benchmarks. However, benchmarks are empirical: they can demonstrate a model's failure on known distributions but cannot guarantee its behavior against unknown inputs. Can we prove that a model will not hallucinate under specific conditions or will not leak sensitive or personal information? This remains a significant open challenge. 

\begin{figure}[tp]
    \centering
    \includegraphics[width=0.8\linewidth]{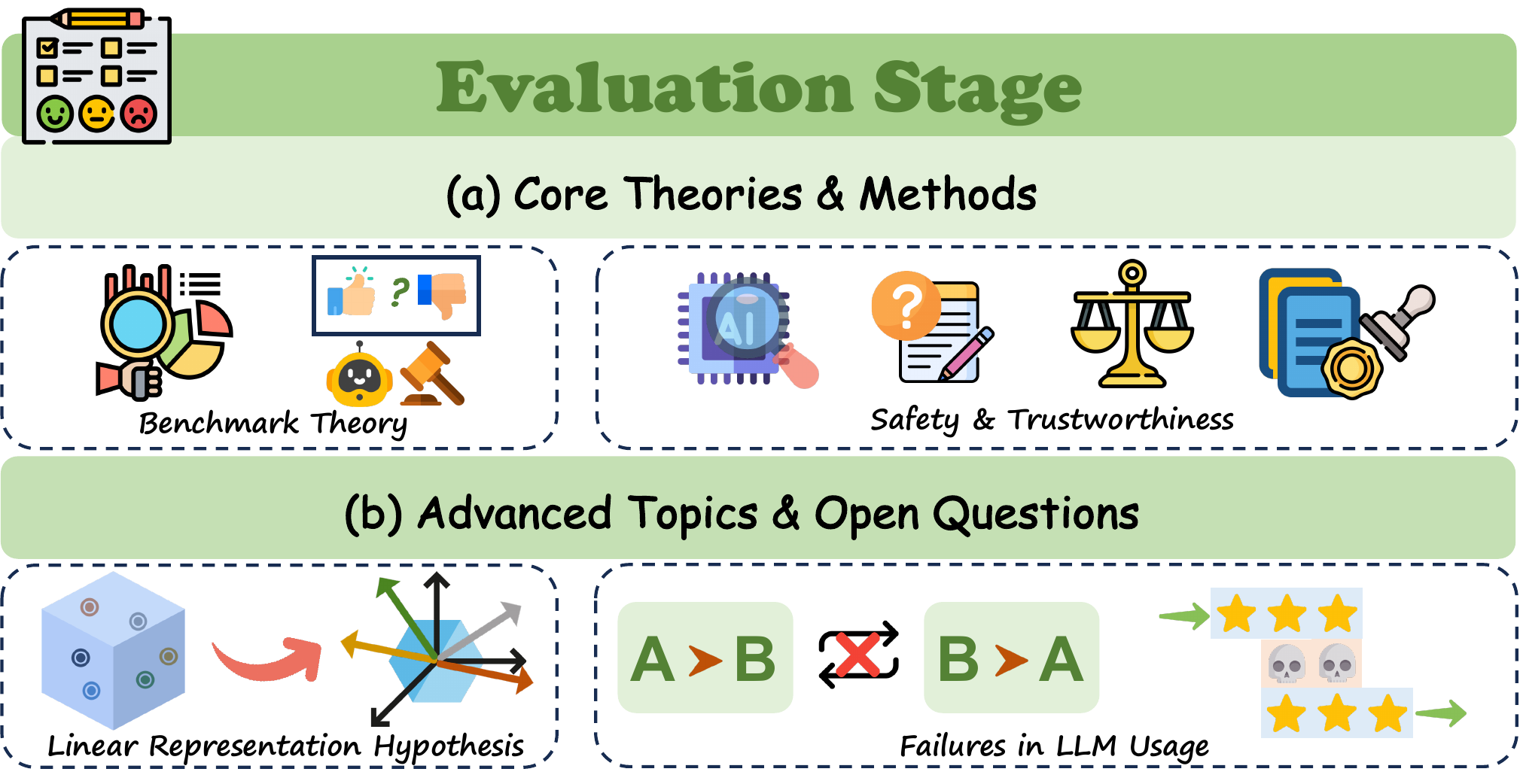}
    \caption{\textbf{An overview of the theoretical landscape in the Evaluation Stage.} This stage is categorized into two dimensions: \textbf{(a) Core Theories \& Methods} focuses on rigorous assessment, including Benchmark Theory (validity and biases of static and judge-based metrics) and Safety \& Trustworthiness (transparency, hallucinations, and formal guarantees). \textbf{(b) Advanced Topics \& Open Questions} highlights frontier challenges, specifically Linear Representation Hypothesis (semantic encoding as linear directions) and Failures in LLM Usage (diagnosing persistent generalization gaps).} 
    \label{fig:evaluation-stage}
\end{figure}

These two fundamental problems define the ultimate theoretical challenges in the Evaluation Stage. In \cref{fig:evaluation-stage} we provide a landscape of the corresponding topics. To begin answering these profound questions, the academic community has initiated several concrete lines of research, each tackling a specific evaluation tool or observable phenomenon. In what follows, we will review these research efforts, detailing how the study of specific strategies 
provides valuable insights. 

\subsection{Core Theories \& Methods}

The fundamental problems define the basics of the entire evaluation stage. In practice, these problems are mainly reflected in specific engineering applications. In the following parts, we will discuss the theoretical advancements that underpinned the implementation of these applications. 
\subsubsection{Benchmark Theory}
To begin answering the profound questions outlined in the fundamental problems, the academic community has initiated concrete lines of research tackling the primary evaluation tools themselves. This work is broadly bifurcated: first, a critical re-examination of the validity of traditional, static benchmarks, and second, a rigorous investigation into the reliability and biases of the emerging ``LLM-as-a-Judge'' paradigm. 

\textbf{Benchmark Validity.} 
A growing body of theoretical and empirical work suggests that high performance on static benchmarks may not correlate with true, generalized capabilities~\citep{zhang2025benchmark}. Several key limitations in current benchmark-based evaluations are identified. 
One primary issue is ``Shortcut Learning''~\citep{du2023shortcut}, where models are found to rely on spurious, non-robust features or biases within a dataset rather than mastering the high-level semantic or reasoning skills the benchmark purports to measure~\citep{chandak2025eliminating}. This leads to a fundamental lack of robustness and exposes that models may be overfitting to the specific artifacts of the test set rather than the underlying task~\citep{lunardi2025robustness}. 
Furthermore, many traditional benchmarks are facing issues of saturation, where top-tier models approach perfect scores, limiting the benchmark's ability to distinguish between SOTA models. To address this, \citet{zhou2025lost} have proposed applying frameworks from psychometrics, such as Item Response Theory, to analyze benchmark quality. This has revealed that many benchmarks suffer from a low ``difficulty ceiling'' and ``item saturation''. \citet{zhang2024inherent} further derive analysis from the perspective of social choice theory, and demonstrate the trade-offs between benchmark diversity and stability. 
Finally, the single-scalar scores produced by most benchmarks obscure the complex combination of skills required for a task. \citet{kim2025benchmark} aim to mechanistically diagnose benchmark composition, decomposing performance into contributions from discrete cognitive abilities. 

\textbf{LLM-as-a-Judge.} 
To overcome the limitations of static benchmarks, especially for evaluating open-ended generation, the ``LLM-as-a-Judge'' (LLM-Judges) paradigm has become widespread~\citep{gu2025surveyllmasajudge}. This approach leverages a powerful LLM to score or rank the outputs of other models. However, this has shifted the theoretical burden from evaluating task performance to evaluating the evaluator itself. 
This paradigm rests on several core assumptions that LLMs can serve as valid human proxies, are capable evaluators, are scalable, and are cost-effective, unfortunately, all of which are being theoretically challenged~\citep{dorner2025limits}. Researchers have argued that LLM-Judges may possess only ``face validity'' rather than true, robust evaluative capacity~\citep{chehbouni2025neither}. 
Two critical flaws being investigated are low reliability and poor psychometric validity. The common practice of using fixed randomness to ensure reproducibility does not guarantee internal consistency. By applying psychometric measures like McDonald's omega and repeating evaluations with different random seeds, studies have found the internal consistency reliability of LLM-Judges to be questionable~\citep{schroeder2024can}. Also, LLM-Judges suffer from severe design flaws that can render their judgments noisy. Key issues include low ``schematic adherence'' and ``factor collapse'' causing the misalignment between the evaluation results and criteria~\citep{feuer2025judgment}. 
Finally, LLM-Judges have been shown to exhibit numerous systematic biases, including position bias, verbosity bias, and authority bias~\citep{ye2024justice,chen-etal-2024-humans}. While some work suggests these biases can be partially mitigated through robust prompting with detailed scoring rubrics~\citep{gao2025evaluating}, the theoretical understanding of these biases remains a critical open area. 

\subsubsection{Safety and Trustworthiness}
Moving beyond the empirical validation provided by benchmarks, we subsequently investigate the foundational theories governing Safety and Trustworthiness, exploring how internal transparency relates to the mathematical boundaries of truthfulness, the explanation for hallucination phenomenon, and the formalization of robustness, fairness, and privacy. 

\textbf{Transparency.} 
Transparency in the context of LLMs refers to the extent to which a model’s internal representations, decision processes, and outputs can be inspected, understood, and communicated to humans, and is often made concrete through methods for model interpretability~\citep{linardatos2020explainable,zhao2024explainability,luo2024understanding,dang2024explainable}. 
Here, we therefore view transparency mainly through the lens of interpretability techniques that aim to reveal how LLMs encode information and produce predictions, and how such insights can support safer and more reliable deployment. Concretely, existing work often groups interpretability methods into three broad categories: global, local, and mechanistic interpretability~\citep{linardatos2020explainable,zhao2024explainability,luo2024understanding}. 
Global interpretability seeks to characterise what a model has learned and how linguistic or semantic information is organised across layers and components. For example, Hewitt and Manning~\citep{hewitt2019structural} introduce structural probes to test whether syntactic dependency trees are encoded as linear structures in contextual word representations. \citet{tenney2019you} use edge-probing tasks to measure how a wide range of linguistic phenomena are distributed across layers in contextual encoders and transformers. 
Local interpretability focuses on explaining individual predictions by attributing them to specific input tokens, features, or intermediate activations. \citet{jain2019attention} show that standard attention weights can be weakly correlated with gradient-based importance scores and thus are not always faithful post-hoc explanations, while \citet{wiegreffe2019attention} argue that, under appropriate definitions and evaluation protocols, attention distributions can still provide useful evidence for explanations. \citet{sundararajan2017axiomatic} propose Integrated Gradients, an axiomatic attribution method that assigns each input feature a contribution score and has become a common tool for token-level importance analysis in neural NLP and LLM outputs. 
Mechanistic interpretability goes a step further by attempting to reverse-engineer specific circuits and features inside LLMs. \citet{elhage2021mathematical} develop a mathematical framework for transformer circuits that characterises the algorithms implemented by small attention-only transformers. \citet{olsson2022context} identify ``induction heads'' as attention heads whose learned algorithm underlies a large fraction of in-context learning. \citet{cunningham2023sparse} use sparse autoencoders to decompose LLM activations into more interpretable latent features, enabling finer-grained localisation and intervention on model behavior. \citet{qian2025demystifying} study the reasoning trajectories of large reasoning models from an information-theoretic perspective, and observe a distinctive ``MI peaks'' phenomenon where the mutual information between intermediate representations and the correct answer suddenly spikes at a few critical generation steps. They further show that these peaks typically align with tokens that express reflection or logical transition, such as ``Hmm'', ``Wait'', or ``Therefore,'' which they term thinking tokens.

\textbf{Hallucination.}
Hallucination refers to instances where an LLM generates outputs that are plausible yet incorrect, conflicting with the model’s world knowledge or context. Recent theoretical research generally yields negative conclusions regarding the complete elimination of hallucinations. \citet{xu2024hallucination} prove that hallucination is mathematically inevitable for any computable LLM, regardless of model architecture or data, due to the inherent limitations of computability and learnability. This inevitability is further corroborated through various theoretical lens, including inductive biases~\citep{wu2024no}, language identification~\citep{kalavasis2025limits}, Bayes-optimal estimators~\citep{liu2025hallucinations}, and calibration~\citep{kalai2024calibrated, kalai2025language}.

Regarding the causes of hallucinations, \citet{zhang2025law} propose the knowledge overshadowing framework, explaining that dominant knowledge suppresses less frequent knowledge during generation. From the perspective of model architecture, \citet{peng2024limitations} argue that transformer architectures have inherent limitations in performing function composition, while \citet{sun2025and} demonstrate that decoder-only transformers act as subsequence embedding models where dominant input subsequences trigger incorrect outputs. Additionally, \citet{kalai2025language} argue that post-training benchmarks exacerbate hallucinations by penalizing uncertainty, effectively incentivizing models to guess rather than abstain.

In terms of mitigation, \citet{kalavasis2025limits} emphasize the role of negative examples, proving that access to negative feedback allows for consistent generation with breadth. \citet{wu2024no} suggest that if facts are restricted to a concept class of finite VC-dimension, non-hallucinating generation is achievable via an improper learner. \citet{zhang2025law} propose amplifying overshadowed knowledge via contrastive decoding to mitigate bias. \citet{kalai2025language} advocate for the modifications of mainstream evaluations to reward appropriate expressions of uncertainty. 
More recently, \citet{liu2025not} characterize three types of uncertainty: document scarcity, limited capability, and query ambiguity, and reveal the fact that current LLMs struggle to accurately identify the root cause and solve it, emphasizing the crucial role of uncertainty in LLM hallucination. 

\textbf{Robustness, fairness and privacy.} 
Ensuring the safe and ethical deployment of LLMs requires addressing critical issues categorized under ``Safety and Trustworthiness''. Among these, robustness, fairness, and privacy are paramount. 
For detailed treatments of these concepts within the LLM domain, we refer to comprehensive surveys (e.g., safety~\citep{shi2024large}, trustworthiness~\citep{huang2024survey,huang2024trustllm,liu2023trustworthy}, fairness~\citep{li2023survey,gallegos2024bias,chu2024fairness}, and privacy~\citep{yao2024survey,yan2024protecting,das2025security}). 

A vast body of literature was dedicated to the theoretical analysis of robustness~\citep{muravev2021certified,ruan2021adversarial}, fairness~\cite{kleinberg2016inherent,liu2019implicit}, and privacy~\citep{li2017differential,kairouz2015composition} in traditional machine learning, primarily because these concepts were often well-defined and could be formalized using precise mathematical objectives and metrics. However, in the current landscape of LLMs, the definitions of robustness, fairness, and privacy can occasionally be more ambiguous, lacking simple closed-form mathematical representations. Furthermore, evaluating these properties often requires using other LLMs as judges or evaluators, which introduces subjectivity and complexity~\citep{guo2023evaluating,li2024llms,gu2024survey}. 
Despite this, there is still some theoretical work to study the related topics within LLMs.
For example, \citet{wolf2023fundamental} introduce a theoretical framework called behavior expectation bounds to formally investigate the fundamental limitations of robustness in LLMs. The core theoretical conclusion, built on this framework, is that for any undesired behavior that an aligned model exhibits with a small, finite probability, there exists an adversarial prompt (whose length increases with the model's complexity) that can trigger this behavior with a probability that approaches one as the prompt length increases. 
This implies a fundamental ``alignment impossibility'': any alignment process that attenuates an undesired behavior but does not remove it entirely (i.e., reduces its probability to a non-zero value) cannot be considered safe against adversarial prompting attacks like jailbreaks~\citep{yi2024jailbreak}. The framework also suggests that popular alignment techniques, such as RLHF, may make the model more susceptible to being prompted into undesired behaviors.

\textbf{Resistance to Misuse.}
The unauthorized or malicious use of LLMs poses significant risks~\citep{liu2023trustworthy}, which severely erode public trust and destabilize information ecosystems. 
To combat these harms, the development and deployment of AI-generated text detection tools~\citep{guo2023close} like watermarking~\citep{kirchenbauer2023watermark} are becoming critical for identifying machine-generated content and ensuring accountability. 
This method allows the output of proprietary LLMs to be algorithmically identified as synthetic with a negligible impact on text quality.
\citet{he2024theoretically} introduce a unified theoretical framework for watermarking LLMs that jointly optimizes both the watermarking scheme and the detector, revealing a fundamental trade-off between watermark detectability (Type-II error) and text distortion. 
\citet{christ2024undetectable} introduce a cryptographically, formally defining it as being computationally infeasible to distinguish watermarked outputs from those of the original model, even with adaptive queries. 
\citet{christ2024provably} prove that the watermark is unremovable under the assumption of adversary uncertainty about the high-quality text distribution, establishing a steep quality degradation versus watermark removal trade-off. 
\citet{hu2023unbiased} introduce the concept of an unbiased watermark for LLMs, which is provably $n$-shot-undetectable, meaning the watermarked output distribution is identical to the original, thereby guaranteeing no degradation in text quality. 
\citet{li2025robust} introduce a method for robust watermark detection called Truncated Goodness-of-Fit test, which models human edits as a sparse mixture distribution problem, and prove Tr-GoF achieves adaptive optimality by reaching the optimal detection boundary of in an asymptotic regime of decreasing watermark signal outperforming existing sum-based methods.
\citet{li2025statistical} introduce a general statistical framework for watermark detection in LLMs based on hypothesis testing using a pivotal statistic, enabling the rigorous evaluation of detection efficiency through class-dependent efficiency (the rate of Type II error decay). 
\citet{hu2024inevitable} propose the two reweight framework and provide a no-go theorem, which proves that it is impossible to simultaneously maintain the highest watermark strength and the highest sampling efficiency when the vocabulary size is greater than two. 

\subsection{Advanced Topics \& Open Questions}
\label{app:open-questions-for-evaluation-stage}
Except for the core theories discussed above, there still remain some open questions for the evaluation stage. These empirically observed phenomena further sparked extensive discussions within the community. In what follows, we will review and discuss the research on these open questions. 

\subsubsection{Linear Representation Hypothesis}

Recent advancements in interpretability have increasingly focused on the Linear Representation Hypothesis (LRH), which posits that high-level semantic concepts are encoded as linear directions within the activation space of LLMs. In this subsection, we review recent theoretical and empirical breakthroughs that formalize, explain, and apply this hypothesis.

Empirical investigations have extensively investigated the emergence of interpretable linear structures within the activation spaces of LLMs. For instance, \citet{gurnee2023language} show that LLMs learn linear representations for spatial and temporal dimensions, effectively mapping geography and history across multiple scales. Similarly, \citet{marks2023geometry} identify a generalized ``truth direction'' within the model's geometry, showing that a simple linear probe can consistently distinguish truthful statements across diverse topics and datasets. \citet{qian2024towards} explore how trustworthiness concepts evolve during the pre-training stage. By applying linear probing technique intermediate checkpoints, they reveal that concepts related to trustworthiness become linearly separable early in the pre-training phase.

Moving beyond observation to theoretical grounding, \citet{jiang2024origins} argue that the interplay between the next-token prediction objective and the implicit bias of gradient descent naturally compels the formation of these linear representations in high-dimensional settings. 
Providing a rigorous geometric framework, \citet{park2023linear} use counterfactual interventions to formalize the LRH in both input and output spaces, and then introduce a ``causal inner product'' that unifies the geometric treatment of linear probing and model steering, thereby giving these directions a clear causal interpretation. 
Furthermore, \citet{marconato2024all} address the universality of these features by establishing an “all-or-none” identifiability theorem, which proves that such linear properties either hold in all or in none of the distributionally equivalent models under specific conditions.
\citet{li2025task} theoretically analyze the efficacy of ``Task Arithmetic,'' proving that, under suitable assumptions, linear operations like addition and negation can successfully edit knowledge in nonlinear Transformers and even generalize to out-of-domain tasks.

\subsubsection{Failures in LLM Usage}

One surprising failure of generalization in auto-regressive LLMs is the reversal curse~\citep{berglund2023reversal}. If a model is trained on a fact in one direction (e.g., ``A is B''), it will fail to automatically generalize to the reverse direction (``B is A''). This means models struggle with basic logical symmetry and exhibit near-zero accuracy when tested on the reversed fact.
Recent theoretical analysis~\citep{zhu2024towards} suggests the reversal curse is a consequence of the asymmetry in the effective model weights learned during the standard (stochastic) gradient descent training process for auto-regressive models. Specifically, the increase of weights for the sequence $\text{A} \to \text{B}$ does not necessarily cause a corresponding increase of weights for the sequence $\text{B} \to \text{A}$. This discovery provides a new theoretical framework and solution direction for understanding and improving the logical reasoning ability of LLMs (including the CoT).

Another famous failure is position bias. 
Position bias in LLMs refers to the tendency of the model to assign disproportionate importance or attention to information based on its location within a long input context. This often manifests as higher weight given to content at the beginning and end of the input.
The specific and more dramatic manifestation of this is the ``Lost-in-the-Middle'' phenomenon~\citep{liu2023lost}, where the model's performance significantly degrades when the crucial, relevant information is placed in the middle of a long input context, even when the model's overall context window is large enough to contain it.
\cite{wu2025on} propose a graph theory framework to analyze position bias in multi-layer Transformers. The research reveals two key insights: causal masking inherently leads to a bias in attention towards the earlier positions of the sequence, as tokens in the deep layers continuously aggregate the context information of earlier tokens; Meanwhile, relative positional encodings introduce the distance attenuation effect to compete and balance with the deviation of the causal mask.


\section{Related Work}
\label{sec:related-work}
LLMs have become a milestone in the development of artificial intelligence. Cutting-edge models are reshaping our paradigm for natural language research~\citep{openai2022introducingchatgpt,guo2025deepseek,bai2023qwen,touvron2023llama,team2023gemini,caruccio2024claude}. These systems have transitioned from specialized tools into general-purpose artifacts capable of human-like reasoning and complex problem-solving. The rapid iteration of these models, driven by massive-scale compute and data, has established a new paradigm in AI development where empirical results often outpace foundational understanding. 

This accelerated development has left the internal operations of LLMs largely opaque, as the sheer scale of trillions of parameters introduces complexities that defy traditional statistical learning intuitions~\citep{scalinglaw1,scalinglaw2}. A primary challenge in the current literature is the emergence of unpredictable behaviors at scale, such as ICL~\citep{brown2020language}, complex hallucinations~\citep{xu2024hallucination}, and the distinct ``aha moments''~\citep{guo2025deepseek} observed during training. While specific studies have pioneered insights into mechanistic interpretability, the existing body of research remains largely fragmented, with theoretical inquiries often isolated from the end-to-end developmental pipeline. 

Consequently, there is an urgent need for a systematic synthesis to transition LLM research from a collection of engineering heuristics toward a principled scientific discipline. This survey contributes to this objective by introducing a unified lifecycle-based taxonomy, identifying the mathematical explanation for LLM phenomena, and the mechanistic origins of emergent intelligence in the next generation of AI systems. 
\section{Conclusion}
\label{sec:conclusion}
In summary, this survey has established a unified lifecycle-based taxonomy to organize the fragmented theoretical landscape of Large Language Models into six critical stages: Data Preparation, Model Preparation, Training, Alignment, Inference, and Evaluation. While LLMs have precipitated a profound paradigm shift in AI through monumental engineering successes, our theoretical understanding of their internal operations remains poor, often forcing us to treat these systems as ``black boxes''. By connecting empirical observations, this work provides a structured roadmap for the community. Ultimately, addressing the identified frontier challenges is essential for transitioning LLM development from a discipline of engineering heuristics toward a principled scientific discipline. 

\bibliography{reference}
\bibliographystyle{tmlr}


\end{document}